\documentclass[10pt]{article} 
\usepackage[preprint]{tmlr}

\usepackage{amsmath,amsfonts,bm}

\def\eqref#1{equation~\ref{#1}}

\def\1{\bm{1}}
\newcommand{\train}{\mathcal{D}}

\DeclareMathAlphabet{\mathsfit}{\encodingdefault}{\sfdefault}{m}{sl}
\SetMathAlphabet{\mathsfit}{bold}{\encodingdefault}{\sfdefault}{bx}{n}

\newcommand{\E}{\mathbb{E}}
\newcommand{\Ls}{\mathcal{L}}

\DeclareMathOperator*{\argmin}{arg\,min}

\usepackage{graphicx}
\usepackage{booktabs}      
\usepackage{multirow}
\usepackage{colortbl}
\usepackage{array}
\usepackage{pifont}
\usepackage{array}
\usepackage{multirow}
\usepackage{longtable}

\usepackage{orcidlink}

\usepackage{url}
\usepackage{hyperref}

\title{T-CLIP: Enabling Thermal Perception for Contrastive Language-Image Pretraining}

\author{\name Tayeba Qazi \email bsz218186@iitd.ac.in \\
      \addr Indian Institute of Technology Delhi, India
      \AND
      \name Ayush Maheshwari \email aymaheshwari@nvidia.com \\
      \addr NVIDIA AI Technology Center, India
      \AND
      \name Prerana Mukherjee \email prerana@jnu.ac.in \\
      \addr Jawaharlal Nehru University, India
      \AND
      \name Brejesh Lall \email brejesh@ee.iitd.ac.in \\
      \addr Indian Institute of Technology Delhi, India
      }

\def\month{MM}  
\def\year{YYYY} 
\def\openreview{\url{https://openreview.net/forum?id=XXXX}} 

\begin{document}

\maketitle

\begin{abstract}
Thermal imaging offers a powerful alternative to visible-spectrum vision under challenging conditions such as low illumination and adverse weather, yet foundational vision-language models like CLIP fail to align thermal images with textual descriptions due to a fundamental thermal perception gap. We identify three major challenges: the lack of captioned thermal datasets, the inability of standard LLMs to reason about thermal phenomena, and a key representational challenge in thermal imaging where global scene context and object-level heat signatures conflict when learned together in a single embedding space. To address these, we introduce IR-Cap, the first physics-aware thermal captioning pipeline and dataset providing complementary global and fine-grained thermal descriptions across three public benchmarks, and T-CLIP, a decoupled dual-LoRA framework that independently adapts CLIP for scene-level and object-level thermal understanding. T-CLIP achieves consistent improvements over all baselines across three thermal benchmarks in cross-modal retrieval, and we provide an exploratory demonstration of its applicability to text-conditioned thermal image generation.
\end{abstract}

\section{Introduction}

\label{sec:intro}

Deep learning has revolutionized computer vision, yet the field remains 
constrained by its heavy reliance on visible-spectrum data. This 
dependency on RGB imagery becomes a failure point in challenging conditions 
such as low-light environments, fog, smoke, or other visual obstructions. 
Thermal imaging, which captures infrared radiation emitted by 
objects, offers a powerful alternative~\citep{vollmer2018infrared}. By 
visualizing heat signatures rather than reflected light, thermal cameras 
operate effectively in total darkness and penetrate various obscurants, 
making them indispensable for applications like 24/7 autonomous navigation, 
search-and-rescue operations, medical diagnostics, and surveillance 
systems~\citep{farooq2023role, wilson2023recent, he2021infrared, 
gade2014thermal, nguyen2021review, hirsh2012hybrid}. Despite these 
advantages, the full potential of thermal imaging remains largely untapped, 
as vision models lack the capability to interpret thermal data.

This limitation is particularly evident in Vision-Language Models (VLMs) such as CLIP \citet{clip}. We identify a significant \textit{thermal perception gap} in CLIP’s understanding, likely stemming from its exclusive training on RGB image-text pairs. Unlike conventional visible-spectrum imagery, where appearance is determined by color and texture, thermal representations are governed by physical properties such as temperature, emissivity, and radiative balance that are largely  absent in CLIP’s pretraining distribution.  
Our analysis reveals that this discrepancy extends beyond a superficial domain shift, manifesting as specific representational limitations.  For instance, when guiding generative models like SDXL \citet{rombach2022high, podell2023sdxl} with prompts such as \textit{``an infrared thermal image of a road with lane markings,"}, the outputs often resemble pseudo-colored RGB images, 
lacking the characteristic thermal signatures expected 
in infrared images. This occurs because the CLIP text encoder relies on optical descriptors like ``white'' and ``yellow'' for lane markings, failing to account for the fact that these features are visible in thermal images due to material emissivity rather than reflected light. Similarly, while CLIP vision encoder can recognize objects via shape contours, it is often insensitive to the underlying heat signatures, which is the primary discriminative signal in the thermal domain. Consequently, the model struggles to differentiate between physically different thermal states, such as a ``warm, running engine'' versus a ``cool, parked car.''

\begin{figure}[t]
  \centering
  \begin{minipage}{0.4\linewidth}
    \includegraphics[width=\linewidth]{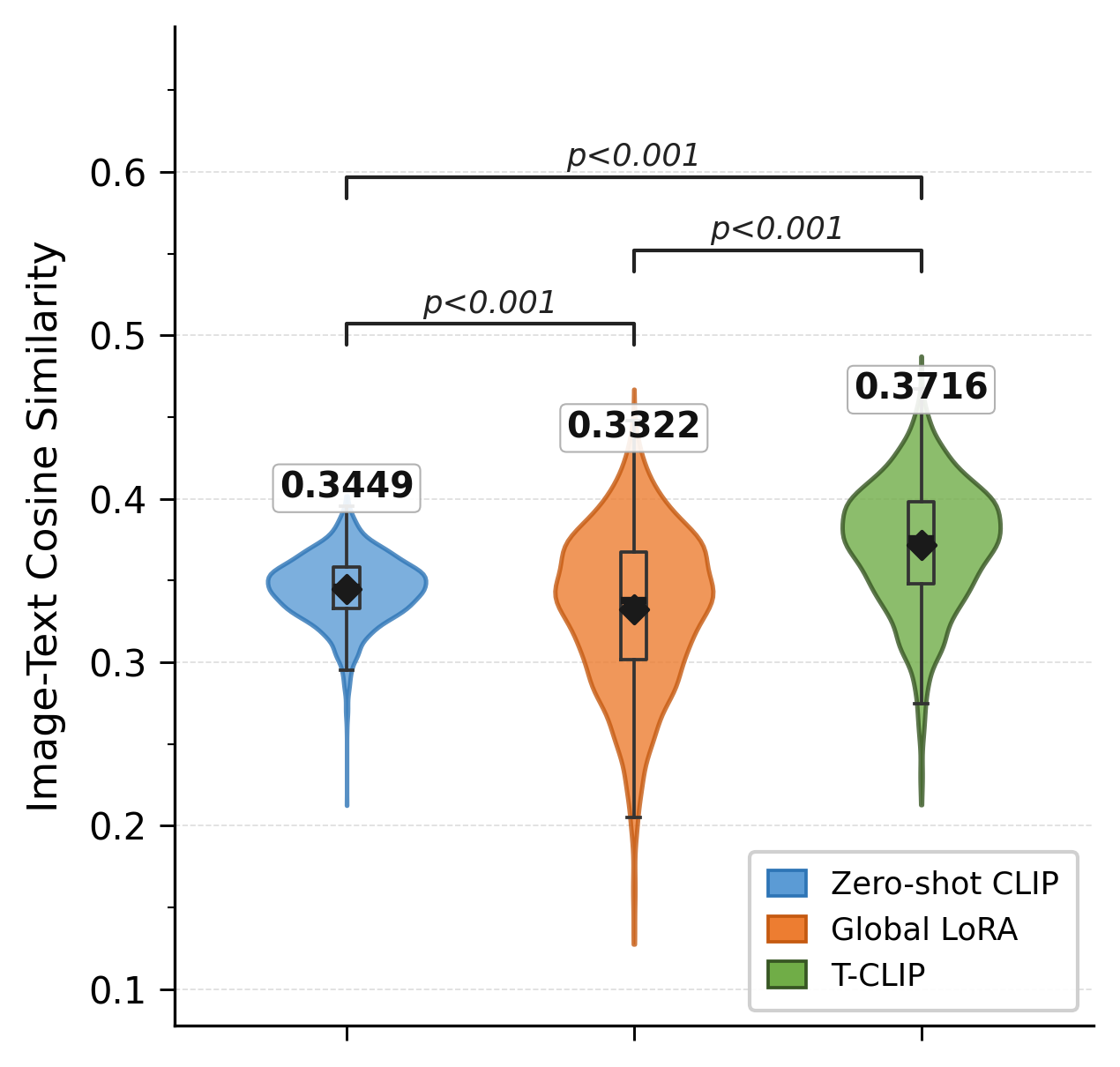}
     \end{minipage}
  \hfill
  \begin{minipage}{0.55\linewidth}
    \fbox{
      \begin{minipage}{0.95\linewidth}
        \small
        Image-text cosine similarity between matched
        thermal image-caption pairs measures how well
        each model aligns a thermal image with its
        correct description in embedding space.
        Zero-shot CLIP (cosine similarity $= 0.3449$)
        shows insufficient thermal image-text alignment,
        resulting in negligible retrieval performance
        (R@1\,$= 0.003$).
        Standard LoRA fine-tuning of CLIP on thermal
        data (Global LoRA) shows a marginal drop in
        cosine similarity ($0.3449 \rightarrow 0.3322$)
        compared to zero-shot CLIP, reflecting
        instability when adapting a model trained on
        RGB data to the thermal domain, using only
        generic scene descriptions without explicit
        thermal supervision.
        T-CLIP, through decoupled dual-context
        alignment on physics-aware captions, achieves
        the strongest alignment ($0.3716$),
        corresponding to a $26\times$ improvement
        in R@1 ($0.003 \rightarrow 0.078$;
        Table~\ref{tab:ECCV_retrieval_kaist}).
        All pairwise differences are statistically
        significant ($p \ll 0.001$, Welch $t$-test,
        $n=2252$); $p$-values are unchanged after
        Benjamini-Hochberg correction
        (FDR\,$=$\,0.05), confirming robustness to
        multiple comparison adjustment
        (appendix section~\ref{sec:supp_cosine}).
      \end{minipage}
    }
  \end{minipage}
  \caption{Mean image-text cosine similarity of
  matched thermal image-caption pairs on the KAIST
  test set~\citep{hwang2015multispectral} ($n = 2252$), reflecting
  the thermal perception gap in standard
  CLIP~\citep{clip} and T-CLIP's improvement.}
  \label{fig:embeddings}
\end{figure}

We quantify this thermal perception gap by measuring the mean
image-text cosine similarity between matched thermal image-caption
pairs. As shown in Figure~\ref{fig:embeddings}, zero-shot CLIP achieves
a cosine similarity of merely 0.3449 on the
KAIST~\citep{hwang2015multispectral} test set (2252 image-text pairs), confirming that
thermal images and their descriptions are poorly aligned in
CLIP's embedding space. This directly translates to negligible
retrieval performance (R@1~$= 0.003$,
Table~\ref{tab:ECCV_retrieval_kaist}), indicating that CLIP's
limitations on thermal data go beyond a simple domain shift,
pointing to a deeper cross-modal representation gap in the
thermal domain. All pairwise cosine similarity differences are
statistically significant ($p \ll 0.001$, Welch $t$-test,
$n=2252$, BH-corrected; appendix section~\ref{sec:supp_cosine}).

Bridging this thermal perception gap requires addressing three key challenges. First, the thermal vision community lacks large-scale captioned datasets; existing datasets are primarily designed for detection or segmentation and lack descriptive textual annotations. Even the IR-500 \citep{ran2025diffv2ir} dataset assigns a generic caption---\textit{``An infrared thermal image''}, to all samples, omitting the rich thermal physics underlying the imagery. Second, standard LLMs fail to generate thermally informative descriptions as they default to RGB-style priors. This bias causes them to hallucinate color and texture attributes while overlooking essential thermal features such as heat signatures, material emissivity, and temperature relationships. Third, our experiments reveal an important insight into the nature of thermal representations. Thermal images carry two distinct levels of information---global-level scene content and object-level heat signatures, that cannot be effectively mapped within a single CLIP embedding space. We demonstrate that simultaneous optimization leads to representational interference, while sequential adaptation results in catastrophic forgetting (Tables~\ref{tab:Different_Captions} and~\ref{tab:Lora_models}). 

To address these challenges, we propose T-CLIP, a simple 
yet effective framework that bridges the thermal perception gap 
through two primary technical contributions. First, we 
develop IR-Cap, a physics-aware thermal captioning pipeline 
that leverages paired RGB images as semantic anchors, enabling 
Qwen2.5-VL-72B-Instruct~\citep{wu2025qwen} to reason about the 
underlying thermal properties of a scene that cannot be directly 
inferred from thermal images by standard LLMs. Using 
specialized dual prompts, we guide the model to generate two 
complementary caption types: one capturing the global scene 
context and the other focusing on object-level thermal attributes 
such as material emissivity, heat retention, and active heat 
signatures. This approach circumvents the inherent inability of 
standard LLMs to interpret thermal phenomena, producing descriptions 
that are both semantically rich and thermally relevant. 
Second, to address the representational challenges specific to 
thermal data, we propose a decoupled dual LoRA framework 
that independently trains two specialized branches, one 
dedicated to scene-level thermal context and another to 
object-level heat signature understanding, on their respective 
caption types, and fuses their complementary representations at 
inference via a weighted ensemble. This 
decoupled design consistently outperforms single-branch variants 
across retrieval metrics and datasets, demonstrating that 
scene-level and object-level thermal representations capture 
complementary information that benefits from specialized 
rather than joint optimization.

Our contributions can be summarized as follows:

\begin{itemize}
\item We introduce \textbf{IR-Cap}, the first thermal caption 
dataset providing physics-aware descriptions beyond generic 
modality labels. Using paired RGB images as semantic anchors, 
we develop a dual-prompt captioning pipeline that generates 
two complementary caption types: Global Thermal Captions 
capturing scene-level context, and Fine-Grained Thermal 
Captions encoding object-level heat signatures. The pipeline applies to any 
paired RGB-thermal dataset; in this work we annotate KAIST, 
FLIR, and FMB, releasing both annotations and pipeline publicly.

\item We present \textbf{T-CLIP}, a decoupled dual-LoRA framework 
for thermal CLIP adaptation. Through controlled ablations and feature 
space analysis, we show that scene-level thermal 
context and object-level heat signatures are geometrically divergent 
and interfere under joint optimization, motivating a decoupled design that leverages IR-Cap's complementary dual captions.

\item T-CLIP achieves consistent improvements over all baselines 
across three thermal benchmarks in image-to-text and text-to-image 
retrieval. Beyond retrieval, we provide an exploratory demonstration showing that the thermally adapted text encoder can be integrated into SDXL for text-conditioned thermal image generation, motivating future systematic study.
\end{itemize}

\section{Related Works}
\label{sec:relatedworks}

\paragraph{CLIP.} Contrastive Language-Image Pre-training (CLIP)~\citep{clip} has established itself as a foundational Vision-Language Model, learning a unified embedding space from extensive RGB image-text pairs. Its zero-shot capabilities have enabled widespread adoption in detection~\citep{vild, glip}, segmentation~\citep{groupvit, lseg}, and generative modeling~\citep{DALLE2, clipdraw}. To avoid the cost of full retraining, lightweight adaptation methods have emerged: CoOp~\citep{yao2023visual} optimizes soft prompt tokens for classification, while CLIP-Adapter~\citep{gao2024clip} introduces bottleneck layers for efficient domain transfer. Despite these advances, CLIP's representations remain intrinsically tied to RGB-based visual semantics, lacking the physical grounding required for non-visible modalities. While specialized adaptations exist for domains such as medical imaging~\citep{lai2023clipath}, the specific challenge of bridging the thermal perception gap remains unaddressed, leaving foundational VLMs insensitive to the underlying physics of thermal images.

\paragraph{Vision-Language Datasets.} The progress in multimodal learning has been fueled by large-scale, captioned datasets in the visible spectrum, such as COCO \citet{COCO}, Visual Genome \citet{VG}, and LAION \citet{laion}. In stark contrast, the thermal imaging community suffers from a severe scarcity of analogous resources. Existing thermal datasets are primarily designed for tasks like classification and object detection, offering only categorical labels (e.g., "person," "car") \citep{hwang2015multispectral,flir,liu2023multi}. Datasets that do include text often provide only generic, modality-focused descriptions, such as "an infrared image" (e.g., IR-500 \citep{ran2025diffv2ir}), which fail to capture the rich, physics-based attributes like heat signatures and material emissivity. This absence of a large-scale, physics-aware thermal caption dataset is a primary bottleneck for developing thermal VLMs, which IR-Cap addresses.

\paragraph{Thermal Image Generation.} Most literature on thermal image synthesis formulates the task as an image-to-image translation problem, converting RGB frames to infrared equivalents~\citep{iwashita2023munet, kniaz2017thermalnet, qazi2025thermaldiff}. Text-conditioned generation remains largely unexplored; while DiffV2IR~\citep{ran2025diffv2ir} provides a notable exception, it relies on generic captions and auxiliary spatial conditioning (RGB and segmentation maps) at inference. This necessitates intensive end-to-end training and limits the model's ability to reason about thermal physics from text alone. In contrast, T-CLIP offers a proof-of-concept for a thermal generation approach only from text prompts without any additional spatial priors.

\begin{figure}[h]
\begin{center}
\includegraphics[width=\linewidth]{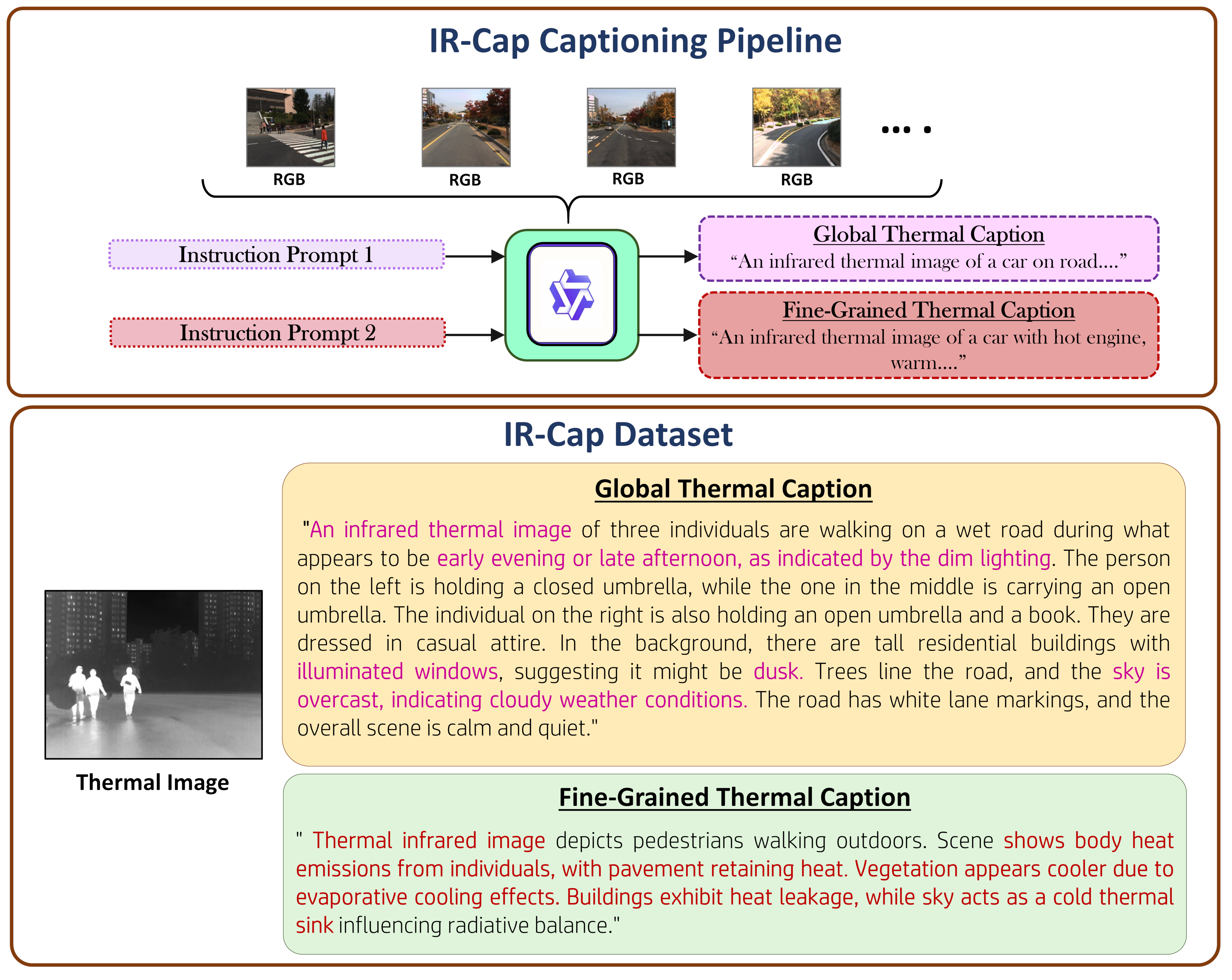}
\end{center}
\caption{IR-Cap captioning pipeline and dataset.
(Top): The IR-Cap captioning pipeline leverages
paired RGB images as semantic anchors to generate
complementary physics-aware captions for thermal images.
Since standard vision-language models are trained on
RGB data and exhibit limited understanding of infrared
thermal characteristics, we condition Qwen2.5-VL-72B-Instruct
on the RGB counterpart of each thermal image and use two
specialized instruction prompts to steer the model toward
reasoning about thermal attributes.
Instruction Prompt~1 elicits a global thermal caption
capturing scene-level environmental context, while
Instruction Prompt~2 elicits a fine-grained thermal caption
encoding object-level heat signatures and thermal phenomena.
(Bottom): A representative IR-Cap dataset sample
showing a thermal image alongside its two complementary
physics-aware captions generated by the pipeline.}
\label{fig:captions}
\end{figure}

\section{IR-Cap Dataset}
\label{sec:ircap}

While conventional RGB images can be described through direct visual 
attributes such as color and texture, thermal appearances are governed 
by physical properties such as temperature, emissivity, and 
environmental context, which are not directly observable in the 
visible spectrum. This semantic-physical gap has limited the 
development of effective thermal-visual reasoning systems. To address 
this, we introduce IR-Cap (Figure~\ref{fig:captions}), the first 
thermal caption dataset providing physics-aware descriptions 
paired with thermal images across three public benchmarks.

\paragraph{Motivation and Captioning Strategy.}
Since standard LLMs cannot reliably interpret 
infrared imagery, we exploit the paired visible-spectrum images 
available in existing multispectral datasets as semantic context. 
Rather than presenting the thermal image directly to the model, 
we condition Qwen2.5-VL-72B-Instruct~\citep{wu2025qwen} on its 
RGB counterpart, allowing it to anchor scene semantics, like object 
identities, spatial layout, and environmental context, while 
our specialized dual prompts redirect its reasoning toward the 
thermal implications of the observed scene, rather than its 
visible-spectrum appearance.

We interpret thermal imagery as comprising
two complementary semantic levels: (i) global scene-level understanding
and (ii) localized object-centric thermal signatures. We observe that
scene-level attributes, including environmental context, weather conditions,
and coarse activity semantics, can often be inferred from aligned RGB imagery
and described reliably by modern vision-language models trained predominantly
on visible-spectrum data. In contrast, fine-grained thermal phenomena, such as
heat intensity, emissivity variations, engine activity, and relative temperature
distributions, are not directly observable from RGB appearance alone.
Consequently, standard vision-language models lack the intrinsic thermal priors
required to accurately characterize such infrared-specific properties.
To address this limitation, we design specialized prompts that encourage the model
to reason about plausible thermal behavior at the object level, enabling the generation
of thermally descriptive annotations, such as distinguishing between the hot engine
region of a moving vehicle and the comparatively cooler engine of a parked vehicle.
These complementary descriptions collectively serve as physics-aware 
semantic priors that bridge the gap between visible-spectrum 
vision-language models and the thermal domain. This perspective inspires our dual prompting strategy, 
described below.
\paragraph{Dual Prompting Strategy.}
We design two distinct instruction prompts, each targeting a 
different level of thermal understanding:

\textbf{Instruction Prompt 1 (Global Thermal Captions, $c_g$):} 
This prompt instructs the model to describe the scene from a 
macro-level thermal perspective, focusing on environmental factors 
that govern the overall heat distribution. Specifically, it asks 
the model to reason about illumination conditions, time of day, 
weather patterns, and material composition comprising of factors that 
collectively determine the thermal landscape of the scene. 
For example, a night scene with wet roads would prompt descriptions 
of reduced ambient heat and increased thermal contrast between 
warm objects and cool surfaces.

\textbf{Instruction Prompt 2 (Fine-Grained Thermal Captions, $c_{fg}$):}
This prompt directs the model toward object-level thermal reasoning, 
focusing on heat signatures, material emissivity differences, and 
temperature relationships between specific objects. For example, 
for the same night scene, this prompt elicits descriptions of 
individual objects' thermal behavior. For example, ``moving vehicles exhibit 
hot engines and warm tires, while pedestrians emit body heat'' , 
capturing the discriminative thermal attributes not present in the  
global descriptions.

Both prompts include the keywords ``infrared'' and ``thermal'' 
in the caption prefix to provide a modality-specific tag, 
ensuring the generated descriptions are framed within the 
thermal domain rather than defaulting to RGB-style language. The full prompt templates are provided in 
the appendix section~\ref{sec:supp_dataset}.
\begin{figure}[h]
\begin{center}
\includegraphics[width=0.50\linewidth]{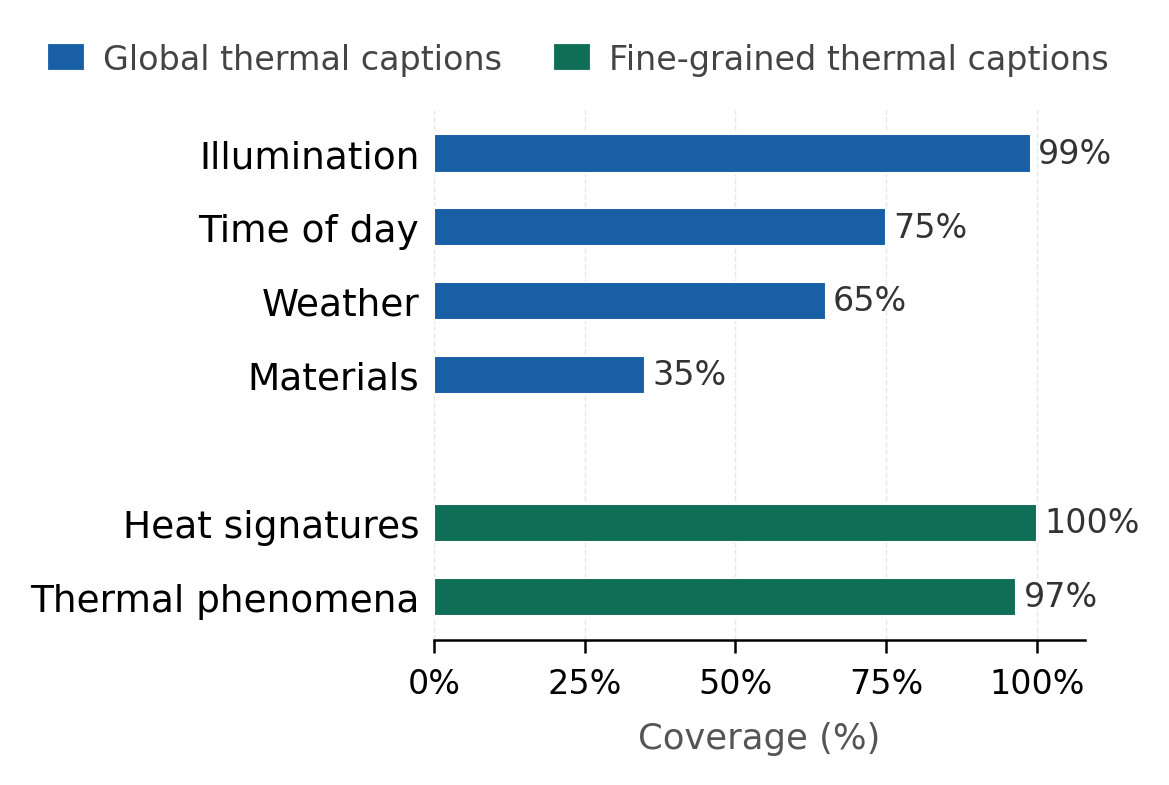}
\end{center}
\caption{Attribute coverage of IR-Cap captions. Values indicate
the percentage of captions within each type that mention the
corresponding attribute.}
\label{fig:ircap_stats}
\end{figure}
\paragraph{Dataset Statistics}

Using this pipeline, we annotate three public thermal
benchmarks, KAIST~\citep{hwang2015multispectral},
FLIR~\citep{flir}, and FMB~\citep{liu2023multi},
augmenting each with paired ($c_g, c_{fg}$)
physics-aware captions.
To quantify attribute coverage, we perform
keyword-based analysis over the full caption corpus,
checking for the presence of predefined attribute
categories (illumination, time of day, weather,
materials for Global captions; heat signatures and
thermal phenomena for Fine-Grained captions) and
reporting the fraction of captions that mention
each attribute.
As illustrated in Figure~\ref{fig:ircap_stats}, this
analysis confirms the broad and complementary
coverage of our dual-prompting strategy.
\textit{Global Thermal Captions} provide necessary
context regarding ambient environmental conditions,
with near-universal coverage of illumination (99\%)
and significant inclusion of temporal and
meteorological context (time of day: 75\%,
weather: 65\%).
In contrast, \textit{Fine-Grained Thermal Captions}
capture intrinsic thermal properties exclusively:
100\% of these captions identify heat signatures,
while 97\% articulate complex phenomena such as
emissivity variations and radiative balance.
Annotating thermal images with such physics-aware metadata provides rich thermal priors that enable comprehensive scene-level contextual understanding and fine-grained object-level reasoning, thereby serving as an essential supervisory signal for mitigating the thermal perception gap identified in section~\ref{sec:challenges}.
\section{Method}
\label{sec:method}

We begin in section~\ref{sec:challenges} by characterizing two fundamental 
challenges in thermal vision-language alignment: the thermal perception 
gap in CLIP (Figure~\ref{fig:embeddings}), and the empirically revealed geometric 
divergence between global and fine-grained thermal feature spaces 
(Figure~\ref{fig:featgeo}), which together motivate our design. Section~\ref{sec:formulation} formalizes the problem, 
after which section~\ref{sec:training} presents our decoupled dual-LoRA training 
strategy, illustrated in Figure~\ref{fig:train}. Finally, 
Figure~\ref{sec:inference} describes the weighted feature fusion at inference (Figure~\ref{fig:inference}).

\subsection{Challenges in Thermal Vision-Language Alignment}
\label{sec:challenges}
\paragraph{Thermal Perception Gap.}
Standard CLIP, pre-trained on RGB-centric web
data, exhibits a significant semantic gap when
applied to thermal imagery. As shown in
Figure~\ref{fig:embeddings}, zero-shot CLIP achieves
a mean image-text cosine similarity of only
$0.3449$ on the KAIST~\citep{hwang2015multispectral}
thermal dataset, despite being paired with
carefully curated thermal captions. This is not
merely a distributional shift; thermal images
encode physically unique signals (emissivity,
heat flux, material conductivity) that are
entirely absent from CLIP's training distribution.
Strikingly, single-LoRA adaptation on global
captions (\textit{Global LoRA}, mean $0.3322$)
degrades alignment relative to zero-shot CLIP,
and all pairwise differences are statistically
significant ($p \ll 0.001$, Welch $t$-test,
BH-corrected; appendix section~\ref{sec:supp_cosine}),
confirming that na\"{i}ve fine-tuning strategies
are insufficient. Effective thermal alignment
requires a purpose-built representation strategy.
\begin{figure}[h]
\begin{center}
\includegraphics[width=0.65\linewidth]{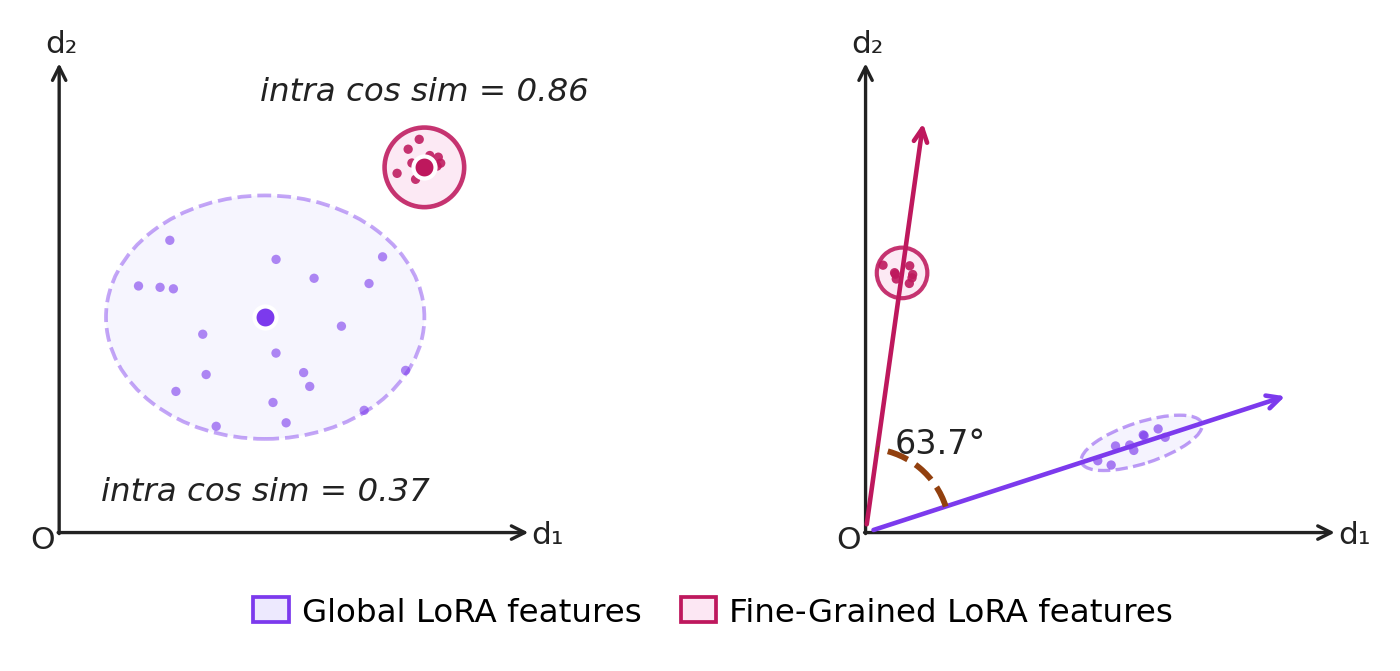}
\end{center}
\caption{Feature geometry of independently trained global and
fine-grained LoRA branches on
KAIST~\citep{hwang2015multispectral}.
(Left) Global LoRA features exhibit low intra-class cohesion
(intra cosine similarity $= 0.37$), while fine-grained LoRA
features form tight compact clusters ($= 0.86$).
(Right) Mean feature vectors subtend $63.7^{\circ}$, confirming
geometric divergence between the two representation spaces and
motivating the decoupled dual-LoRA design
(see section~\ref{sec:challenges}).}
\label{fig:featgeo}
\end{figure}
\paragraph{Complementary yet Divergent Thermal Semantics.}
Thermal imagery encodes information at two conceptually distinct semantic 
levels: \textit{global scene characteristics}, encompassing environmental 
conditions, time of day, spatial layout, and material composition, which 
partially overlap with visible-spectrum cues, and \textit{object-level 
thermal physics}, including emissivity variations, localized heat signatures, 
and thermal conductivity differences, which are exclusively observable in the 
thermal domain and demand physics-aware representation. We empirically reveal 
these two levels to be not only semantically complementary but geometrically 
divergent in feature space, by training two LoRA branches independently — one exclusively on global 
scene captions and the other exclusively on fine-grained thermal captions and probing their respective feature distributions as visualized in 
Figure~\ref{fig:featgeo}, global LoRA features exhibit low intra-class cohesion 
(mean intra cosine similarity $= 0.37$), reflecting the inherent diversity 
of scene-level thermal contexts, while fine-grained LoRA features form tight, 
compact clusters (mean intra cosine similarity $= 0.86$), consistent with the 
physically constrained nature of object-level heat signatures. Furthermore, 
the mean feature vectors of the two branches subtend an angle of 
$63.7^{\circ}$, confirming that global and fine-grained thermal 
representations occupy significantly different regions of the embedding space. Joint optimization of a 
single LoRA on mixed captions conflates these divergent objectives, producing 
gradient interference that degrades both, as shown in 
Table~\ref{tab:Different_Captions}. Sequential training partially mitigates 
interference but leaves the second-stage LoRA without awareness of the 
first-stage objectives, leading to suboptimal joint representations, as 
confirmed by Table~\ref{tab:Lora_models}. These findings motivate a fully 
decoupled strategy with independent parameter spaces, which we describe next.

\subsection{Problem Formulation}
\label{sec:formulation}

Given a thermal image $I_{\text{th}} \in \mathbb{R}^{H \times W \times 1}$ and two paired captions — a global thermal caption $c_g$ describing environmental context and a fine-grained caption $c_{fg}$ encoding object-specific heat signatures, we aim to learn, over paired-caption dataset $\mathcal{D}$, complementary 
representation spaces that capture both semantic levels without interference. We introduce two independent Low-Rank Adapter (LoRA) modules $\theta_g$ and $\theta_f$, each applied to both the vision and text encoders of a frozen CLIP~\citep{clip} backbone, and optimize them separately:

\begin{equation}
\label{eq:global_lora}
    \theta_g^* = \argmin_{\theta_g} \;
    \E_{(I_{\text{th}}, c_g) \sim \train}
    \left[ \Ls_{\text{global}}\!\left(F_{\theta_g}(I_{\text{th}}),\,
    G_{\theta_g}(c_g)\right) \right]
\end{equation}
\begin{equation}
\label{eq:fg_lora}
    \theta_f^* = \argmin_{\theta_f} \;
    \E_{(I_{\text{th}}, c_{fg}) \sim \train}
    \left[ \Ls_{\text{fg}}\!\left(F_{\theta_f}(I_{\text{th}}),\,
    G_{\theta_f}(c_{fg})\right) \right]
\end{equation}
where $F_{\theta_g}(\cdot) = F_{\text{frozen}}(\cdot) +
\Delta F_{\theta_g^v}(\cdot)$ and
$G_{\theta_g}(\cdot) = G_{\text{frozen}}(\cdot) +
\Delta G_{\theta_g^t}(\cdot)$ denote the base encoders augmented
with their respective LoRA adaptations, and analogously for $\theta_f$.
Since the two modules are optimized independently over disjoint
supervisory signals, with the shared CLIP backbone kept frozen
throughout, there is zero gradient interference between the two
representation spaces.

Beyond the geometric motivation established in section~\ref{sec:challenges}, LoRA 
adaptation is particularly suited to the thermal domain, where large-scale paired datasets are scarce; full fine-tuning of CLIP 
would therefore risk catastrophic forgetting of its rich visual-semantic priors given the limited thermal 
training data, whereas LoRA confines adaptation to a small set of low-rank 
parameters while preserving the frozen backbone's generalizable 
representations.

To our 
knowledge, T-CLIP is the first framework to explicitly decompose thermal 
image understanding into global scene context and object-level heat 
signature physics, empirically reveal their geometric divergence in feature 
space (Figure~\ref{fig:featgeo}), and leverage this finding as an architectural 
prior instantiated through independent LoRA parameter spaces on a shared 
frozen CLIP backbone, with ablations confirming the necessity of this 
decoupling (Tables~\ref{tab:Different_Captions} and~\ref{tab:Lora_models}).

\begin{figure}[h]
\begin{center}
\includegraphics[width=\linewidth]{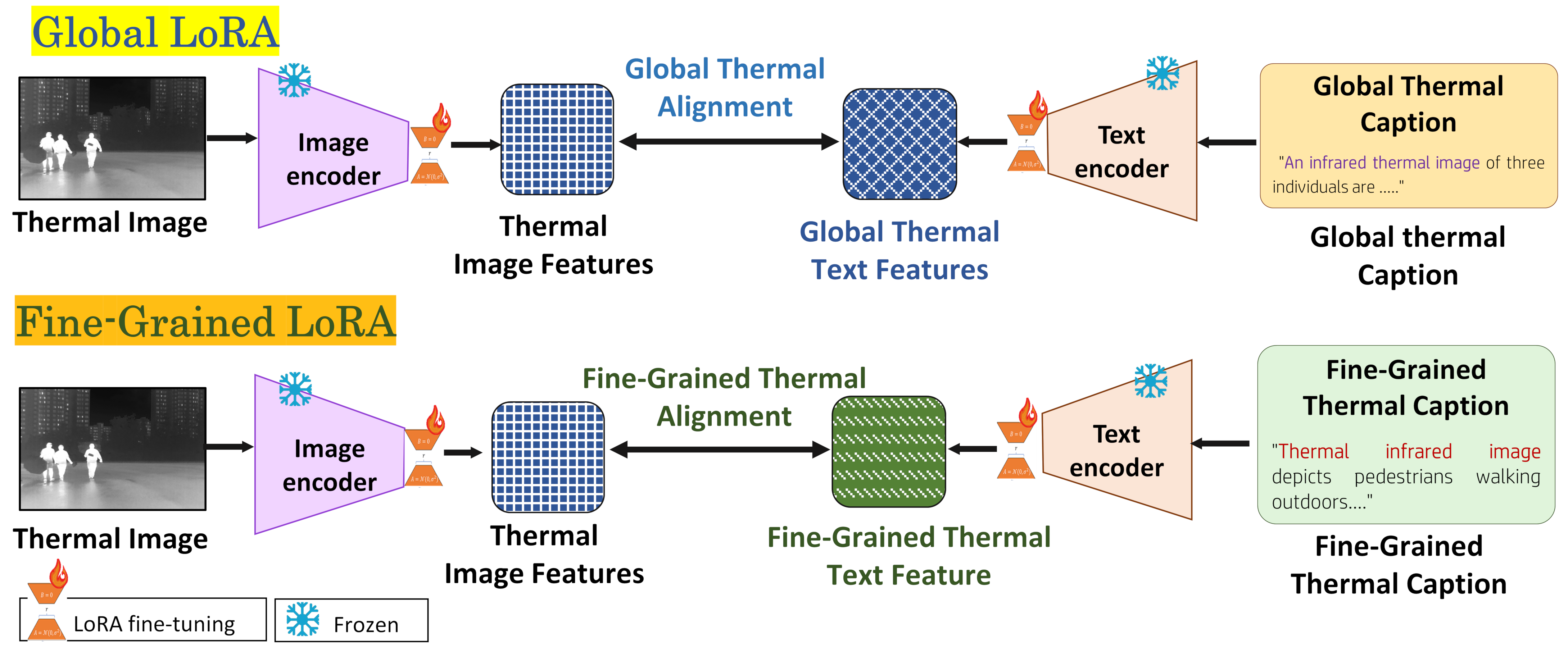}
\end{center}
\caption{T-CLIP dual LoRA training pipeline. Two independent LoRA
modules on a frozen CLIP backbone are optimized on semantically
distinct caption types, preventing gradient interference while
enabling complementary thermal representation learning.}
\label{fig:train}
\end{figure}

\subsection{Dual LoRA Training for Decoupled Thermal Representation}
\label{sec:training}

The overall training pipeline is illustrated in Figure~\ref{fig:train}, with the corresponding pseudocode provided in section~\ref{sec:supp_pseudocode} of the appendix. Each LoRA branch is trained independently with a standard InfoNCE 
contrastive objective~\citep{clip}, operating on the same set of thermal 
images but with semantically distinct caption types.

\paragraph{Global Thermal LoRA.}
$\theta_g$ is trained exclusively on global captions $c_g$ describing
scene content, illumination, weather, material composition, and
temporal context:
\begin{equation}
\label{eq:global}
    \mathcal{L}_{\text{global}}(\theta_g) =
    -\log \frac{
        \exp\!\left(\mathrm{sim}(F_{\theta_g}(I_{\text{th}}),\,
        G_{\theta_g}(c_g))\,/\,\tau\right)
    }{
        \sum_{k=1}^{B} \exp\!\left(\mathrm{sim}(F_{\theta_g}(I_{\text{th}}),\,
        G_{\theta_g}(c_g^k))\,/\,\tau\right)
    }
\end{equation}
where $\mathrm{sim}(\cdot,\cdot)$ denotes cosine similarity, $\tau$ is
a learned temperature parameter, $B$ is the batch size, and $c_g^k$
denotes the $k$-th global caption in the batch serving as a negative
sample. The same notation applies to $c_{fg}^k$, (the $k$-th
fine-grained caption) in $\mathcal{L}_{\text{fg}}$ (Equation~\ref{eq:fg}).
This enforces $\theta_g$ to encode the macroscopic thermal landscape ---
spatial heat flow, environmental gradients, and scene-level structure,
without contamination from object-level supervisory signal.

\paragraph{Fine-Grained Thermal LoRA.}
$\theta_f$ is trained exclusively on $c_{fg}$, captions describing
localized thermal phenomena such as emissivity contrasts, heat source
attribution, and object-specific temperature anomalies (\textit{e.g.},
``hot engine block,'' ``warm pedestrian torso''):
\begin{equation}
\label{eq:fg}
    \mathcal{L}_{\text{fg}}(\theta_f) =
    -\log \frac{
        \exp\!\left(\mathrm{sim}(F_{\theta_f}(I_{\text{th}}),\,
        G_{\theta_f}(c_{fg}))\,/\,\tau\right)
    }{
        \sum_{k=1}^{B} \exp\!\left(\mathrm{sim}(F_{\theta_f}(I_{\text{th}}),\,
        G_{\theta_f}(c_{fg}^k))\,/\,\tau\right)
    }
\end{equation}

The decoupled weight spaces preserve the geometric divergence observed in Figure~\ref{fig:featgeo} rather than collapsing it, and gradients from neither branch interfere with the other throughout training.
\begin{figure}[h]
\begin{center}
\includegraphics[width=\linewidth]{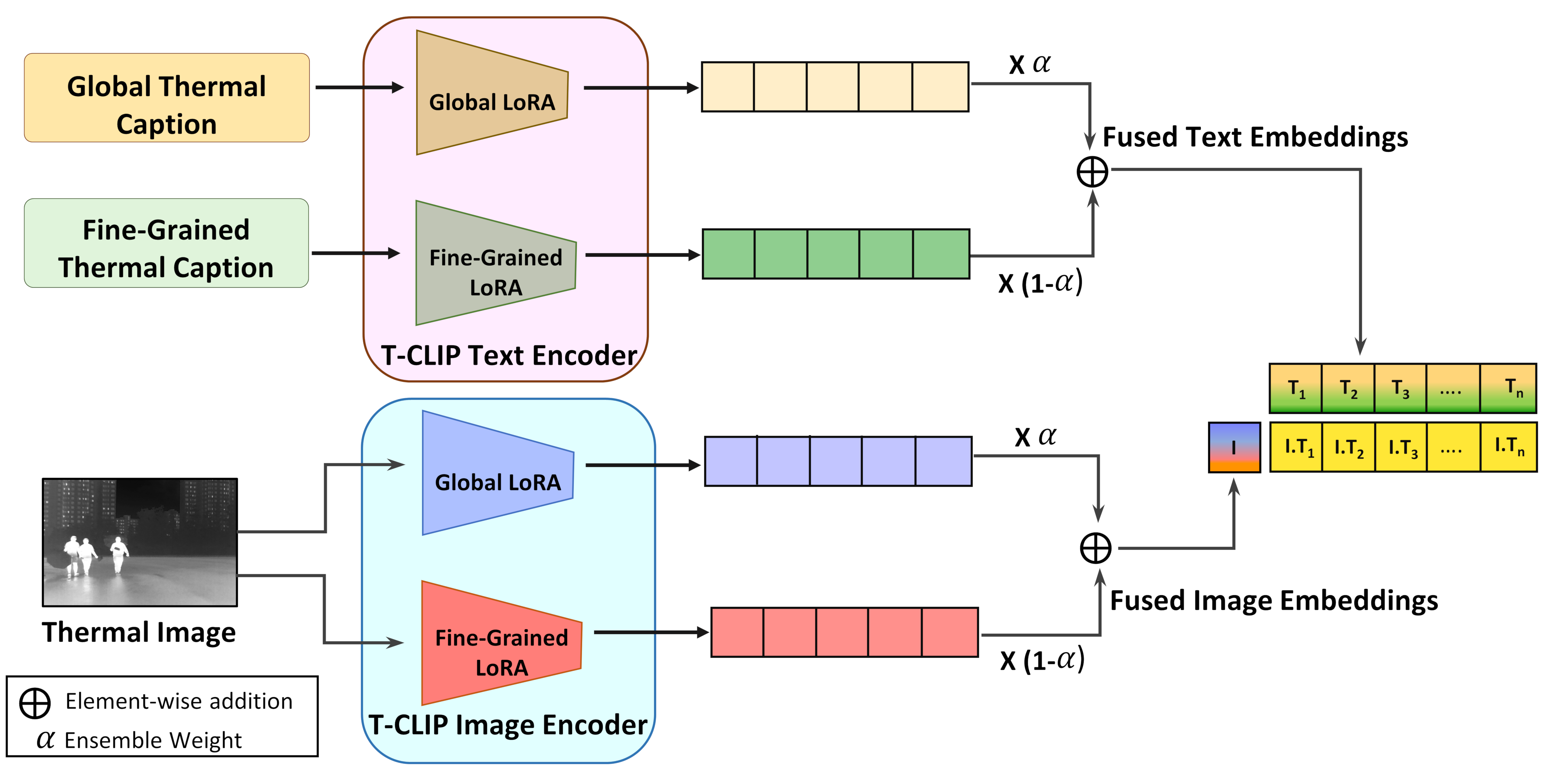}
\end{center}
\caption{T-CLIP inference. Features from the global and fine-grained
LoRA branches are combined via a scalar hyperparameter $\alpha$,
enabling flexible control over scene-level versus object-level
emphasis at retrieval time.}
\label{fig:inference}
\end{figure}
\subsection{Inference via Weighted Feature Fusion}
\label{sec:inference}
Figure~\ref{fig:inference} illustrates the inference pipeline. At test time,
both LoRA branches independently process a thermal image and their
outputs are fused via a scalar hyperparameter $\alpha$:
\begin{equation}
\label{eq:image_fusion}
    v_{\text{fused}} = \alpha \cdot F_{\theta_g}(I_{\text{th}})
                     + (1 - \alpha) \cdot F_{\theta_f}(I_{\text{th}})
\end{equation}
For text-side fusion during text-to-image retrieval, captions from
both branches are combined symmetrically:
\begin{equation}
\label{eq:text_fusion}
    w_{\text{fused}} = \alpha \cdot G_{\theta_g}(c_g)
                     + (1 - \alpha) \cdot G_{\theta_f}(c_{fg})
\end{equation}
$\alpha \in [0,1]$ controls the relative contribution of each branch;
we set $\alpha = 0.8$ by default, a value determined empirically
(Figure~\ref{fig:alpha_ablation}), reflecting that global thermal context
provides the dominant discriminative signal while fine-grained features
supply complementary specificity.
This fusion requires only a weighted sum of two forward passes, adding
negligible computational overhead at inference relative to standard
CLIP~\citep{clip}.
\section{Experiments}
\label{sec:experiments}

We evaluate T-CLIP on thermal image-text retrieval 
(section~\ref{sec:retrieval_results}) as our primary task and further explore its thermal image generation capability 
(section~\ref{sec:imggen}).

\subsection{Experimental Settings}
\label{sec:exp_settings}

\paragraph{Evaluation Datasets.}
We evaluate on three publically available thermal imaging benchmarks. \textbf{KAIST}~\citep{hwang2015multispectral} is a large-scale 
multispectral pedestrian dataset from which we use 50,000 unique 
thermal images spanning diverse outdoor and urban driving scenarios 
across both day and night conditions. \textbf{FLIR}~\citep{flir} is a driving-focused 
thermal dataset; following~\citep{zhang2020multispectral}, we use the 
aligned version with 4,128 training and 1,013 validation pairs. \textbf{FMB}~\citep{liu2023multi} is a multimodal benchmark augmented 
with 12,200 thermal images following the procedure 
of~\citep{mayr2024narrowing}. All results are reported on standard test 
splits to ensure fair comparison across methods.

For each dataset, global thermal captions and fine-grained thermal 
captions were generated using our proposed IR-Cap captioning pipeline 
(described in section~\ref{sec:ircap}), which produces paired global scene-level and 
object-level thermal descriptions for every image. 
\paragraph{Evaluation Settings.}
We report Recall@K (R@K) for $K \in \{1, 5, 10, 25, 50, 100\}$, 
measuring the percentage of queries for which the correct match appears 
within the top-$K$ retrieved results. We evaluate in both 
image-to-text (I2T) and text-to-image (T2I) directions across all three 
datasets. All evaluation protocols are held constant across methods 
and datasets. For each method and dataset, we report the mean and standard deviation of R@K over three independent training runs with different random seeds (0, 42, 123).

\paragraph{Training Settings.}
T-CLIP is fine-tuned for 10,000 steps with a batch size of 64 on a 
single NVIDIA A6000 GPU, requiring approximately 1.5 hours of training, highlighting the practical viability of our lightweight LoRA-based 
approach. The two LoRA branches ($\theta_g$ and $\theta_f$) are trained 
independently on their respective caption types, with all 
hyperparameters held constant across both training runs. Detailed 
configurations, including learning rate, LoRA rank, and optimizer 
settings, are listed in appendix section~\ref{sec:supp_hyperparams}.

\subsection{Comparing with Baselines}
\label{sec:retrieval_results}

We compare T-CLIP against the following baselines, all evaluated on 
captions generated by our IR-Cap pipeline to ensure a fair and 
controlled comparison.

\paragraph{Zero-shot CLIP.} We evaluate the off-the-shelf 
CLIP~\citep{clip} model without any thermal adaptation, serving as a 
domain-agnostic lower bound. Since IR-Cap generates two caption types, 
we report two zero-shot CLIP variants: \textit{CLIP (Global)}, evaluated 
on global scene captions, and \textit{CLIP (F-G)}, evaluated on 
fine-grained thermal captions.

\paragraph{CLIP-Adapter.} We include two CLIP-Adapter~\citep{gao2024clip} 
variants: \textit{CLIP-Adapter (Global)}, trained and evaluated on global 
thermal captions, and \textit{CLIP-Adapter (F-G)}, trained and evaluated on 
fine-grained thermal captions. Both learn a lightweight residual feature 
adapter on top of frozen CLIP features without modifying the encoder weights, 
serving as an alternative parameter-efficient adaptation baseline.

\paragraph{DeCLIP.} Full fine-tuning of CLIP on small thermal datasets risks catastrophic forgetting of pretrained visual-semantic representations. As a representative full-parameter baseline, we therefore include DeCLIP~\citep{declip}, a data-efficient extension that trains all CLIP parameters using image--image and text--text nearest-neighbour consistency via a momentum encoder.  We report two variants: \textit{DeCLIP (Global)}, trained on global thermal captions, and \textit{DeCLIP (F-G)}, trained on fine-grained thermal captions.

\paragraph{Single LoRA Baselines.} We fine-tune two independent 
single-branch LoRA models on the IR-Cap dataset: \textit{Global LoRA}, 
fine-tuned and evaluated exclusively on global scene captions, and 
\textit{Fine-Grained LoRA}, fine-tuned and evaluated exclusively on 
fine-grained thermal captions. These isolate the contribution of each 
caption type and directly motivate our dual-branch design.

\paragraph{T-CLIP Variants.}We report three inference variants of T-CLIP, all using the fused 
dual-LoRA representation with $\alpha = 0.8$. In \textit{T-CLIP (Global)}, a single global caption is passed through 
both LoRA branches at inference, with the resulting branch features 
fused to produce the final representation. \textit{T-CLIP (F-G)} 
follows the same protocol, substituting the global caption with a 
fine-grained thermal caption; 
and in \textit{T-CLIP (Dual)}, our proposed model, the Global LoRA branch independently processes the global caption while the Fine-Grained LoRA branch processes the fine-grained caption, with both fused to obtain the final embedding.
In scenarios where only a single caption type is available, 
\textit{T-CLIP (Global)} and \textit{T-CLIP (F-G)} serve as natural 
fallbacks; the available caption is routed through both branches 
without any architectural modification, preserving the dual-branch 
fusion mechanism and ensuring applicability across standard image-text 
retrieval settings where paired captions are not provided a priori.

Results are reported in Tables~\ref{tab:ECCV_retrieval_kaist},~\ref{tab:ECCV_retrieval_flir}, and~\ref{tab:ECCV_retrieval_fmb};
detailed mean\,$\pm$\,std across all recall thresholds are provided
in section~\ref{sec:supp_retrieval} of the appendix.

Across all three datasets, zero-shot CLIP \citep{clip} yields consistently poor retrieval performance, confirming that standard RGB-trained representations are fundamentally inadequate for thermal image-text alignment irrespective of whether global or fine-grained captions are used as queries.

Parameter-efficient adaptation substantially improves over zero-shot CLIP, with CLIP-Adapter \citep{gao2024clip}, DeCLIP \citep{declip}, and single-branch LoRA baselines all showing clear gains. DeCLIP achieves intermediate performance, generally surpassing CLIP-Adapter but falling short of LoRA-based methods across datasets. Global LoRA consistently and significantly outperforms both CLIP-Adapter and DeCLIP across all metrics and datasets, indicating that for thermal domain alignment, encoder weight modification via LoRA is more effective than residual feature adaptation or data-efficient self-supervision. Among single-branch methods, Global LoRA outperforms Fine-Grained LoRA across all datasets, suggesting that scene-level context provides a stronger retrieval signal in isolation, consistent with the feature geometry analysis in Figure~\ref{fig:featgeo}.

T-CLIP (Dual) outperforms all baselines across the majority of metrics and datasets, demonstrating that the complementary thermal semantics captured by the two decoupled LoRA branches are mutually beneficial at retrieval time. On KAIST (Table~\ref{tab:ECCV_retrieval_kaist}), T-CLIP (Dual) achieves relative R@1 gains of +10.5\% (I2T) and +21.9\% (T2I) over Global LoRA, the strongest single-branch baseline, with consistent improvements on FLIR (Table~\ref{tab:ECCV_retrieval_flir}) (+17.1\% I2T, +10.4\% T2I). Importantly, comparing T-CLIP (Global) and T-CLIP (F-G) against their single-branch LoRA counterparts reveals that the dual-LoRA representation enriches the visual embedding beyond what either caption type achieves independently. This confirms that training on complementary thermal semantics improves the quality of the learned representation space itself, not merely the inference-time fusion.

On FMB (Table~\ref{tab:ECCV_retrieval_fmb}), T-CLIP (Global) marginally outperforms T-CLIP (Dual) at lower I2T recall thresholds (R@1--R@10), though both variants consistently surpass Global LoRA baseline in this range. T-CLIP (Dual) recovers the lead at R@25 and above for I2T and dominates across all T2I metrics, consistent with the trend observed on KAIST and FLIR.

\begin{table}[t]
  \caption{Text-image retrieval recall@K on the KAIST~\citep{hwang2015multispectral}
  test set. R@1 reported as mean\,$\pm$\,std across three independent training
  runs; all other R@K are means. \textbf{Bold}: best result.
  \underline{Underline}: second best. F-G = fine-grained.
  $\Delta$: relative R@K gain of T-CLIP (Dual) over the strongest baseline
  (Global LoRA).}
  \label{tab:ECCV_retrieval_kaist}
\begin{center}
  \resizebox{\linewidth}{!}{%
  \begin{tabular}{l  @{\hspace{2.0em}} c c c c c c @{\hspace{2.5em}} c c c c c c}
  & \multicolumn{6}{c}{\bf IMAGE-TO-TEXT RETRIEVAL} &
    \multicolumn{6}{c}{\bf TEXT-TO-IMAGE RETRIEVAL} \\
  \bf METHOD & \bf R@1 & \bf \bf R@5 & \bf R@10 & \bf R@25 & \bf R@50 & \bf R@100
         & \bf R@1 & \bf R@5 & \bf R@10 & \bf R@25 & \bf R@50 & \bf R@100 
\\ \hline \\
  CLIP (Global)
    & 0.003 & 0.016 & 0.034 & 0.068 & 0.110 & 0.179
    & 0.002 & 0.013 & 0.027 & 0.048 & 0.080 & 0.135 \\
  CLIP (F-G)
    & 0.001 & 0.004 & 0.008 & 0.029 & 0.057 & 0.096
    & 0.001 & 0.007 & 0.014 & 0.030 & 0.048 & 0.088 \\
  CLIP-Adapter (Global)
    & $0.018{\pm}0.001$ & 0.072 & 0.123 & 0.229 & 0.357 & 0.521
    & $0.020{\pm}0.002$ & 0.070 & 0.119 & 0.234 & 0.360 & 0.538 \\
  CLIP-Adapter (F-G)
    & $0.009{\pm}0.002$ & 0.041 & 0.081 & 0.167 & 0.283 & 0.436
    & $0.010{\pm}0.000$ & 0.046 & 0.082 & 0.166 & 0.281 & 0.445 \\
  DeCLIP (Global)
    & $0.038{\pm}0.002$ & 0.129 & 0.206 & 0.349 & 0.497 & 0.644
    & $0.041{\pm}0.000$ & 0.133 & 0.209 & 0.357 & 0.496 & 0.650 \\
  DeCLIP (F-G)
    & $0.013{\pm}0.000$ & 0.048 & 0.088 & 0.169 & 0.273 & 0.433
    & $0.014{\pm}0.000$ & 0.056 & 0.094 & 0.175 & 0.285 & 0.439 \\
  Global LoRA
    & $0.070{\pm}0.002$ & 0.224 & 0.332 & 0.520 & 0.685 & 0.825
    & $0.069{\pm}0.005$ & 0.221 & 0.331 & 0.521 & 0.679 & 0.822 \\
  F-G LoRA
    & $0.021{\pm}0.002$ & 0.073 & 0.125 & 0.248 & 0.381 & 0.571
    & $0.019{\pm}0.002$ & 0.079 & 0.139 & 0.260 & 0.400 & 0.582 \\
  T-CLIP (Global)
    & $\underline{0.071{\pm}0.003}$ & \underline{0.233} & \underline{0.349}
    & \underline{0.541} & \underline{0.695} & \underline{0.837}
    & $\underline{0.078{\pm}0.002}$ & \underline{0.242} & \underline{0.359}
    & \underline{0.554} & \underline{0.710} & \underline{0.843} \\
  T-CLIP (F-G)
    & $0.016{\pm}0.002$ & 0.064 & 0.101 & 0.194 & 0.311 & 0.473
    & $0.012{\pm}0.002$ & 0.048 & 0.084 & 0.157 & 0.255 & 0.394 \\
  \textbf{T-CLIP (Dual)}
    & $\mathbf{0.078{\pm}0.004}$ & \textbf{0.252} & \textbf{0.374}
    & \textbf{0.571} & \textbf{0.729} & \textbf{0.862}
    & $\mathbf{0.084{\pm}0.002}$ & \textbf{0.264} & \textbf{0.386}
    & \textbf{0.580} & \textbf{0.740} & \textbf{0.870} \\
  \rowcolor{blue!8}
  $\Delta$ vs.\ Global LoRA
    & +10.5\% & +12.4\% & +12.7\% & +9.7\% & +6.3\% & +4.5\%
    & +21.9\% & +19.4\% & +16.8\% & +11.3\% & +9.1\% & +5.8\% \\
  \end{tabular}
  }
  \end{center}
\end{table}

\begin{table}[t]
  \caption{Text-image retrieval recall@K on the FLIR~\citep{flir}
  test set. \textbf{Bold}: best result. \underline{Underline}: second best.
  F-G\,$=$\,fine-grained. $\Delta$: relative R@K gain of T-CLIP (Dual) over
  the strongest baseline (Global LoRA).
  R@1 reported as mean\,$\pm$\,std across three independent training runs.}
    \label{tab:ECCV_retrieval_flir}
 \begin{center}
  \resizebox{\linewidth}{!}{%
  \begin{tabular}{l  @{\hspace{2.0em}} c c c c c c @{\hspace{2.5em}} c c c c c c}
  & \multicolumn{6}{c}{\bf IMAGE-TO-TEXT RETRIEVAL} &
    \multicolumn{6}{c}{\bf TEXT-TO-IMAGE RETRIEVAL} \\
  \bf METHOD & \bf R@1 & \bf \bf R@5 & \bf R@10 & \bf R@25 & \bf R@50 & \bf R@100
         & \bf R@1 & \bf R@5 & \bf R@10 & \bf R@25 & \bf R@50 & \bf R@100 
\\ \hline \\

  CLIP (Global)
    & 0.014 & 0.070 & 0.115 & 0.242 & 0.360 & 0.513
    & 0.014 & 0.062 & 0.106 & 0.229 & 0.319 & 0.467 \\
  CLIP (F-G)
    & 0.006 & 0.022 & 0.039 & 0.110 & 0.220 & 0.332
    & 0.005 & 0.019 & 0.032 & 0.093 & 0.158 & 0.253 \\

  CLIP-Adapter (Global)
    & $0.045{\pm}0.006$ & 0.153 & 0.247 & 0.411 & 0.545 & 0.700
    & $0.031{\pm}0.003$ & 0.130 & 0.217 & 0.381 & 0.529 & 0.706 \\
  CLIP-Adapter (F-G)
    & $0.027{\pm}0.003$ & 0.081 & 0.134 & 0.250 & 0.402 & 0.582
    & $0.019{\pm}0.001$ & 0.075 & 0.133 & 0.244 & 0.367 & 0.555 \\

  DeCLIP (Global)
    & $0.049{\pm}0.002$ & 0.154 & 0.228 & 0.376 & 0.507 & 0.642
    & $0.045{\pm}0.002$ & 0.158 & 0.232 & 0.367 & 0.510 & 0.641 \\
  DeCLIP (F-G)
    & $0.012{\pm}0.001$ & 0.056 & 0.104 & 0.197 & 0.312 & 0.474
    & $0.015{\pm}0.001$ & 0.073 & 0.114 & 0.218 & 0.324 & 0.474 \\

  Global LoRA
    & $0.105{\pm}0.003$ & 0.329 & 0.487 & 0.678 & 0.816 & 0.914
    & $0.106{\pm}0.003$ & 0.321 & 0.459 & 0.673 & 0.812 & 0.907 \\
  F-G LoRA
    & $0.027{\pm}0.004$ & 0.089 & 0.150 & 0.290 & 0.453 & 0.642
    & $0.035{\pm}0.003$ & 0.127 & 0.206 & 0.369 & 0.528 & 0.702 \\

  T-CLIP (Global)
    & $\underline{0.119{\pm}0.004}$ & \underline{0.353}
    & \underline{0.504} & \underline{0.707}
    & \underline{0.833} & \underline{0.923}
    & $\underline{0.105{\pm}0.007}$ & \underline{0.348}
    & \underline{0.492} & \underline{0.712}
    & \underline{0.839} & \underline{0.925} \\
  T-CLIP (F-G)
    & $0.025{\pm}0.005$ & 0.107 & 0.182 & 0.316 & 0.473 & 0.649
    & $0.020{\pm}0.004$ & 0.069 & 0.130 & 0.239 & 0.361 & 0.523 \\
  \textbf{T-CLIP (Dual)}
    & $\mathbf{0.123{\pm}0.010}$ & \textbf{0.357}
    & \textbf{0.517} & \textbf{0.727}
    & \textbf{0.851} & \textbf{0.931}
    & $\mathbf{0.117{\pm}0.003}$ & \textbf{0.364}
    & \textbf{0.511} & \textbf{0.728}
    & \textbf{0.851} & \textbf{0.932} \\
  \rowcolor{blue!8}
  $\Delta$ vs. Global LoRA
    & $+$17.1\% & $+$8.5\% & $+$6.2\%
    & $+$7.2\% & $+$4.3\% & $+$1.9\%
    & $+$10.4\% & $+$13.4\% & $+$11.3\%
    & $+$8.2\% & $+$4.8\% & $+$2.8\% \\
  \end{tabular}
      }
  \end{center}
\end{table}


\begin{table}[t]
\caption{Text-image retrieval recall@K on the FMB~\citep{liu2023multi}
  test set. R@1 reported as mean\,$\pm$\,std across three independent training
  runs; all other R@K are means. \textbf{Bold}: best result.
  \underline{Underline}: second best. F-G = fine-grained.
  $\Delta$: relative R@K gain of the best T-CLIP variant per metric over
  Global LoRA.}
    \label{tab:ECCV_retrieval_fmb}
  \begin{center}
  \resizebox{\linewidth}{!}{%
  \begin{tabular}{l  @{\hspace{2.0em}} c c c c c c @{\hspace{2.5em}} c c c c c c}
  & \multicolumn{6}{c}{\bf IMAGE-TO-TEXT RETRIEVAL} &
    \multicolumn{6}{c}{\bf TEXT-TO-IMAGE RETRIEVAL} \\
  \bf METHOD & \bf R@1 & \bf \bf R@5 & \bf R@10 & \bf R@25 & \bf R@50 & \bf R@100
         & \bf R@1 & \bf R@5 & \bf R@10 & \bf R@25 & \bf R@50 & \bf R@100 
\\ \hline \\
  CLIP (Global)
    & 0.043 & 0.118 & 0.175 & 0.318 & 0.443 & 0.632
    & 0.018 & 0.071 & 0.143 & 0.243 & 0.418 & 0.618 \\
  CLIP (F-G)
    & 0.011 & 0.082 & 0.143 & 0.225 & 0.279 & 0.532
    & 0.004 & 0.032 & 0.068 & 0.157 & 0.289 & 0.479 \\
  CLIP-Adapter (Global)
    & $0.075{\pm}0.007$ & 0.266 & 0.409 & 0.605 & 0.768 & 0.905
    & $0.063{\pm}0.002$ & 0.259 & 0.402 & 0.613 & 0.775 & 0.927 \\
  CLIP-Adapter (F-G)
    & $0.029{\pm}0.004$ & 0.157 & 0.277 & 0.455 & 0.677 & 0.880
    & $0.039{\pm}0.004$ & 0.148 & 0.243 & 0.438 & 0.611 & 0.863 \\
  DeCLIP (Global)
    & $0.095{\pm}0.013$ & 0.268 & 0.400 & 0.573 & 0.704 & 0.834
    & $0.077{\pm}0.005$ & 0.277 & 0.370 & 0.586 & 0.700 & 0.843 \\
  DeCLIP (F-G)
    & $0.041{\pm}0.009$ & 0.145 & 0.238 & 0.402 & 0.561 & 0.743
    & $0.038{\pm}0.013$ & 0.136 & 0.239 & 0.400 & 0.566 & 0.754 \\
  Global LoRA
    & $\underline{0.104{\pm}0.015}$ & \underline{0.342} & 0.496 & 0.731 & 0.866 & 0.956
    & $0.100{\pm}0.012$ & 0.350 & 0.510 & 0.737 & 0.882 & 0.960 \\
  F-G LoRA
    & $0.037{\pm}0.002$ & 0.133 & 0.248 & 0.479 & 0.658 & 0.842
    & $0.048{\pm}0.002$ & 0.185 & 0.282 & 0.489 & 0.666 & 0.848 \\
  T-CLIP (Global)
    & $\mathbf{0.105{\pm}0.012}$ & \textbf{0.362} & \textbf{0.539}
    & \underline{0.762} & \underline{0.871} & \underline{0.958}
    & $\underline{0.101{\pm}0.006}$ & \underline{0.385} & \underline{0.556}
    & \underline{0.787} & \underline{0.901} & \underline{0.970} \\
  T-CLIP (F-G)
    & $0.042{\pm}0.002$ & 0.151 & 0.245 & 0.413 & 0.580 & 0.788
    & $0.039{\pm}0.006$ & 0.144 & 0.219 & 0.346 & 0.526 & 0.708 \\
  \textbf{T-CLIP (Dual)}
    & $0.098{\pm}0.017$ & 0.342 & \underline{0.537}
    & \textbf{0.771} & \textbf{0.895} & \textbf{0.973}
    & $\mathbf{0.104{\pm}0.015}$ & \textbf{0.395} & \textbf{0.587}
    & \textbf{0.818} & \textbf{0.914} & \textbf{0.979} \\
  \rowcolor{blue!8}
  $\Delta$ vs.\ Global LoRA
    & $+$1.0\% & $+$5.8\% & $+$8.7\% & $+$5.5\% & $+$3.3\% & $+$1.8\%
    & $+$4.0\% & $+$12.9\% & $+$15.1\% & $+$11.0\% & $+$3.6\% & $+$2.0\% \\
  \end{tabular}
  }
  \end{center}
\end{table}
\subsection{Ablation Study}
\paragraph{Effect of Caption Type.}
Table~\ref{tab:Different_Captions} ablates the effect of caption type on retrieval performance on KAIST. Global LoRA substantially outperforms Fine-Grained LoRA, confirming that scene-level thermal context is more discriminative than object-level heat signatures in isolation, consistent with the feature geometry analysis in Figure~\ref{fig:featgeo}.
Training a single LoRA on an equal mixture of global and fine-grained captions (Mixed 50:50) reveals that it degrades performance on global captions relative to specialized Global LoRA, yet surpasses specialized F-G LoRA when evaluated on fine-grained captions. This suggests that exposure to global captions during training enriches the representation for fine-grained retrieval, but the inclusion of fine-grained captions simultaneously dilutes the global discriminability. This trade-off demonstrates that the two caption types impose conflicting supervision signals under shared training, motivating our decoupled design in T-CLIP where both caption types are optimized independently rather than under joint supervision.
\begin{table}[t]
\caption{Ablation of caption types on KAIST~\citep{hwang2015multispectral}.
  Global LoRA and fine-grained (F-G) LoRA are trained and evaluated on
  their respective caption types. Mixed (50:50) trains a single LoRA on
  an equal mixture of global and fine-grained captions as a representative
  mixed training baseline; (Global) and (F-G) denote the evaluation
  caption type.}
\label{tab:Different_Captions}
\begin{center}
\resizebox{\linewidth}{!}{%
\begin{tabular}{l @{\hspace{2.0em}} c  c  c  c  c c}
& \multicolumn{3}{c}{\bf IMAGE-TO-TEXT RETRIEVAL} &
  \multicolumn{3}{c}{\bf TEXT-TO-IMAGE RETRIEVAL} \\
\bf CAPTION TYPE & \bf R@1 & \bf R@10 &\bf R@25 & \bf R@1 & \bf R@10 & \bf R@25 \\ \hline \\

Global LoRA
    & \textbf{0.070} & \textbf{0.332} & \textbf{0.520}
    & \textbf{0.069} & \textbf{0.331} & \textbf{0.521} \\
F-G LoRA
    & 0.021 & 0.125 & 0.248
    & 0.019 & 0.139 & 0.260 \\
Mixed (50:50) (Global)
    & 0.054 & 0.318 & 0.496
    & 0.054 & 0.287 & 0.476 \\
Mixed (50:50) (F-G)
    & 0.027 & 0.160 & 0.297
    & 0.021 & 0.152 & 0.283 \\
\end{tabular}
}
\end{center}

\end{table}
\paragraph{Single vs.\ Cross-Initialized vs.\ Dual LoRA.}
Table~\ref{tab:Lora_models} compares single-branch LoRA, cross-initialized LoRA variants, and our decoupled Dual LoRA across different caption evaluation settings.
Dual LoRA with dual-caption evaluation achieves the best performance across all metrics, outperforming all single-branch and initialized variants. Notably, initializing Fine-Grained LoRA from Global LoRA weights leads to performance collapse, suggesting that fine-grained thermal representations cannot be reliably obtained by adapting from global ones and that the two optimization trajectories are incompatible. That Global LoRA also fails when initialized from Fine-Grained LoRA validates the geometric incompatibility of the two representation spaces, reflecting the significant divergence reported in Figure~\ref{fig:featgeo}. Furthermore, Dual LoRA evaluated on global captions (R@1: 0.071) performs comparably to single-branch Global LoRA (R@1: 0.070) while supporting fine-grained retrieval. This capability is entirely absent in any single-branch model, thereby confirming that the dual-branch design retains specialization in each caption domain without requiring separate inference-time models.

\begin{figure}[h]
\begin{center}
\includegraphics[width=0.65\linewidth]{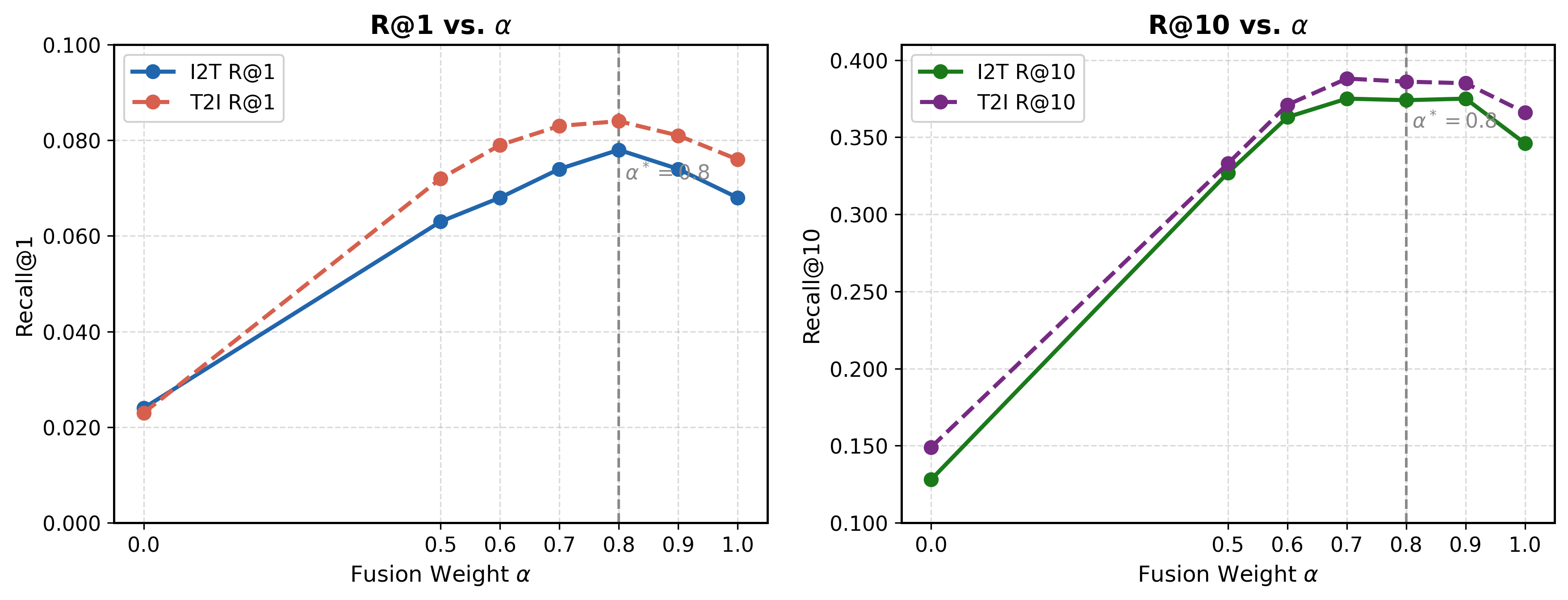}
\end{center}
\caption{Fusion weight $\alpha$ ablation (KAIST~\citep{hwang2015multispectral}).
Peak performance at $\alpha=0.8$ confirms the complementary contribution
of global and fine-grained branches, with steeper degradation toward
$\alpha=0.0$ reflecting the dominance of global thermal context.}
\label{fig:alpha_ablation}
\end{figure}

\begin{table}[t]
   \caption{Ablation of LoRA module designs on KAIST~\citep{hwang2015multispectral}. For dual LoRA, the caption type
  used for evaluation is indicated. A$\vert$B denotes that model B is
  initialized from the weights of model A.}
 \label{tab:Lora_models}
    \begin{center}

  \resizebox{\linewidth}{!}{%
 \begin{tabular}{l @{\hspace{2.0em}} c c c  c c c}
    & \multicolumn{3}{c}{\bf IMAGE-TO-TEXT RETRIEVAL} &
      \multicolumn{3}{c}{\bf TEXT-TO-IMAGE RETRIEVAL} \\

    \bf MODEL & \bf R@1 & \bf R@10 & \bf R@25
               & \bf R@1 & \bf R@10 & \bf R@25 \\
 \hline \\
    \textit{Single LoRA} & & & & & & \\
    \quad Global LoRA 
        & 0.070 & 0.332 & 0.520 
        & 0.069 & 0.331 & 0.521 \\
    \quad F-G LoRA 
        & 0.021 & 0.125 & 0.248 
        & 0.019 & 0.139 & 0.260 \\
    \textit{Initialized} & & & & & & \\
    \quad F-G LoRA$\vert$Global LoRA 
        & 0.002 & 0.013 & 0.028 
        & 0.003 & 0.021 & 0.041 \\
    \quad Global LoRA$\vert$F-G LoRA 
        & 0.008 & 0.044 & 0.087 
        & 0.014 & 0.076 & 0.140 \\
    \textit{Dual LoRA} & & & & & & \\
    \quad Global Captions 
        & 0.071 & 0.349 & 0.541 
        & 0.078 & 0.359 & 0.554 \\
    \quad F-G Captions 
        & 0.016 & 0.101 & 0.194 
        & 0.012 & 0.084 & 0.157 \\
    \rowcolor{blue!8} \quad Dual Captions 
        & \textbf{0.078} & \textbf{0.374} & \textbf{0.571} 
        & \textbf{0.084} & \textbf{0.386} & \textbf{0.580} \\
    \end{tabular}
    }
    \end{center} 
\end{table}
\paragraph{Vision vs. Text Encoder Adaptation.} Table~\ref{tab:encoder_ablation} ablates whether LoRA parameters are applied to the vision encoder only, text encoder only, or both. Adapting a single encoder falls significantly short of adapting both jointly, confirming that effective thermal alignment requires simultaneous adaptation of both modalities.
\begin{table}[t]
  \label{tab:encoder_ablation}
  \caption{Ablation of LoRA fine-tuning scope across encoder
configurations on KAIST~\citep{hwang2015multispectral}. All variants
use dual captions during evaluation, with $\alpha=0.8$.}
    \begin{center}
      \resizebox{\linewidth}{!}{%

    \begin{tabular}{l @{\hspace{2.0em}} c  c  c  c  c  c}
    & \multicolumn{3}{c}{\bf IMAGE-TO-TEXT RETRIEVAL} & 
      \multicolumn{3}{c}{\bf TEXT-TO-IMAGE RETRIEVAL} \\
    \bf FINE-TUNING SCOPE & \bf R@1 & \bf R@10 & \bf R@25 
                               & \bf \bf R@1 & \bf R@10 & \bf R@25 \\
    \hline \\
    Vision Encoder Only        
        & 0.028 & 0.183 & 0.328 
        & 0.028 & 0.185 & 0.318 \\
    Text Encoder Only          
        & 0.016 & 0.131 & 0.237 
        & 0.017 & 0.131 & 0.257 \\
    \rowcolor{blue!8} Both Encoders (T-CLIP Dual)   
        & \textbf{0.078} & \textbf{0.374} & \textbf{0.571} 
        & \textbf{0.084} & \textbf{0.386} & \textbf{0.580} \\
    \end{tabular}
    }
   \end{center}
   
\end{table}
\paragraph{Sensitivity of Retrieval Performance to $\alpha$.}
Figure~\ref{fig:alpha_ablation} shows retrieval performance across
$\alpha \in \{0.0, 0.5, 0.6, 0.7, 0.8, 0.9, 1.0\}$ on KAIST.
Performance peaks at $\alpha = 0.8$ and degrades at both extremes,
with the higher performance at $\alpha = 1.0$ relative to $\alpha = 0.0$
underscoring that global thermal context is the dominant retrieval signal.
Nevertheless, the sharp drop at both extremes validates that both
caption types are essential; fine-grained thermal captions contribute
meaningfully beyond global context alone, and the model benefits from
their complementary fusion.
We empirically verify that $\alpha = 0.8$
is optimal across all three datasets
(KAIST~\citep{hwang2015multispectral},
FLIR~\citep{flir}, and
FMB~\citep{liu2023multi}),
and set it as the default for all
reported results.
\subsection{Parameter and Training Efficiency} \label{sec:efficiency} 

As shown in Table~\ref{tab:efficiency}, T-CLIP (Dual) achieves the best 
retrieval performance with two 
independent 737M LoRA branches which are identical in size to a single Global 
LoRA. Since the two branches are fully independent, their training is 
parallelizable, reducing the wall-clock training time to $\sim$1 hr 
when run simultaneously on two GPUs, which is equivalent to a single Global 
LoRA training run. The performance gain is therefore attributable 
entirely to the decoupled dual-caption design of T-CLIP (Dual) rather than increased 
model capacity, which is a favorable trade-off between architectural complexity 
and retrieval performance.
\begin{table}[t]
   \caption{Efficiency comparison of all methods. Trainable parameters
and wall-clock training time are reported for a single training run
on KAIST on a single NVIDIA A6000 GPU with batch size 128.
$\times$2 denotes two independent training runs for T-CLIP (Dual),
which can be parallelized to $\sim$1\,hr on two GPUs. R@1 results
on KAIST~\citep{hwang2015multispectral} test set.}
      \label{tab:efficiency}
    \begin{center}

      \resizebox{\linewidth}{!}{%

    \begin{tabular}{l  c  c  c  c  c  }
    \bf METHOD & \bf TRAINABLE PARAM. & \bf WALL-CLOCK TIME 
                    & \bf I2T R@1 & \bf T2I R@1 \\
                    \hline \\
    Zero-shot CLIP      & 0              & 0              & 0.003 & 0.002 \\
    CLIP-Adapter        & 131M           & $\sim$40 min   & 0.018 & 0.020 \\
    LoRA (Text Only)    & 294M           & $\sim$1 hr     & 0.016 & 0.017 \\
    LoRA (Vision Only)  & 442M           & $\sim$1 hr     & 0.028 & 0.028 \\
    Global LoRA         & 737M           & $\sim$1 hr     & 0.070 & 0.069 \\
   \rowcolor{blue!8} T-CLIP (Dual) 
        & \textbf{737M $\times$ 2} & \textbf{$\sim$1 hr $\times$ 2} 
        & \textbf{0.078} & \textbf{0.084} \\
    \end{tabular}
    }
   \end{center}
\end{table}
\subsection{Cross-Dataset Generalization}
\label{sec:generalization}
Table~\ref{tab:OOD_results} evaluates T-CLIP (Dual) on the FMB test set when trained on other thermal datasets. Despite the sensor and scene distribution shift, cross-dataset models retain meaningful retrieval capability, with KAIST- and FLIR-trained models recovering 75--85\% of FMB-trained performance at R@1. This demonstrates that T-CLIP learns thermal representations that transfer across sensors and environmental conditions, which is a noteworthy result given the high sensor-dependence of thermal imagery. Training on the combined set further improves performance across all metrics, indicating that diversity of thermal scenes and sensor types is beneficial alongside dataset-specific fine-tuning.
\begin{table}[t]
   \caption{Cross-dataset generalization on the FMB~\citep{liu2023multi} 
test set. Each row denotes the dataset used for training. Combined set 
denotes training on all three datasets, KAIST~\citep{hwang2015multispectral}, 
FLIR~\citep{flir}, and FMB~\citep{liu2023multi}, evaluated on FMB.}
  \label{tab:OOD_results}
    \begin{center}
      \resizebox{\linewidth}{!}{%

    \begin{tabular}{l @{\hspace{2.0em}} c  c  c  c  c  c}
    & \multicolumn{3}{c}{\bf IMAGE-TO-TEXT RETRIEVAL} & \multicolumn{3}{c}{\bf TEXT-TO-IMAGE RETRIEVAL} \\
    \bf TRAINED ON & \bf R@1 & \bf R@10 & \bf  R@25 & \bf R@1 & \bf R@10 & \bf R@25 \\
    \hline \\
    FMB  & 0.098 & 0.537 & 0.771 & 0.104 & 0.587 & 0.818 \\
    KAIST  & 0.079 & 0.421 & 0.671 & 0.082 & 0.496 & 0.668 \\
    FLIR  & 0.075 & 0.375 & 0.561 & 0.111 & 0.439 & 0.604 \\
   \rowcolor{blue!8} Combined Set & \textbf{0.132} & \textbf{0.654} & \textbf{0.850} & \textbf{0.125} & \textbf{0.650} & \textbf{0.864} \\
    \end{tabular}
  }
     \end{center} 
\end{table}
\subsection{IR-Cap Caption Quality Evaluation}
\label{sec:ircap_eval}

We evaluate IR-Cap caption quality through
a human study with five domain-expert
annotators assessing 150 stratified
image-caption pairs per caption type
(300 pairs total), sampled uniformly
across dataset origin
(KAIST~\citep{hwang2015multispectral},
FLIR~\citep{flir}, FMB~\citep{liu2023multi}),
time of day, and scene type.
Annotators rated each caption on four
dimensions for Global captions and five
for Fine-Grained captions:
Thermal Accuracy (D1), Semantic Correctness
(D2), RGB Hallucination (D3, binary),
Specificity (D4), and Emissivity Reasoning
(D5, Fine-Grained only).
A caption is approved if it meets minimum
thresholds on all applicable dimensions
simultaneously.
Full protocol details, annotator assignment
design, rubric anchors, and failure case
analysis are provided in section~\ref{sec:supp_ircap} of the
appendix.

Table~\ref{tab:ircap_eval} presents the results.
Global Thermal Captions achieve an overall
approval rate of $97.3\%$ $[94.1, 99.0]$,
while Fine-Grained Thermal Captions achieve
$84.0\%$ $[77.8, 89.0]$ (95\% CI).
All dimensions achieve Krippendorff's
$\alpha > 0.60$ (substantial agreement),
with RGB Hallucination reaching
$\alpha > 0.80$, consistent with its
binary and objective nature.
Global
captions score significantly higher on
Thermal Accuracy (D1: $3.58$ vs.\ $3.21$,
paired $t$-test $p{<}0.001$), indicating
that scene-level thermal properties are
more reliably inferred from RGB anchors
than object-level heat signatures.
The primary limiting factor for Fine-Grained
captions is Emissivity Reasoning
(D5: $89.3\%$ pass rate), reflecting
object-level thermal physics are
not completely estimated from visible-spectrum
appearance, a limitation of the
RGB-anchoring strategy. Failure case analysis is provided in
appendix section~\ref{sec:supp_failures}.

\begin{table}[t]

\caption{IR-Cap human evaluation results
($n{=}150$ per caption type,
stratified across
KAIST~\citep{hwang2015multispectral},
FLIR~\citep{flir}, and
FMB~\citep{liu2023multi}).
Scores are mean\,$\pm$\,std on the
1--4 Likert scale (higher is better);
D3 is hallucination-free percentage.
Appr.\% denotes the percentage of
captions meeting the per-dimension
pass threshold.
Overall approval requires passing all
applicable dimensions simultaneously;
95\% CIs via Wilson score interval.
$^\dagger$Fine-grained only. \protect \footnotemark}
\label{tab:ircap_eval}
\begin{center}

\begin{tabular}{l  c c  c c}
& \multicolumn{2}{c}{\bf GLOBAL}
& \multicolumn{2}{c}{\bf FINE-GRAINED} \\
\bf DIMENSION
  & \bf SCORE & \bf APPR.\%
  & \bf SCORE & \bf APPR.\% \\
\hline \\
D1: Thermal Accuracy
  & $3.58\pm0.52$ & $98.7\%$
  & $3.21\pm0.68$ & $91.3\%$ \\
D2: Semantic Correctness
  & $3.74\pm0.48$ & $99.3\%$
  & $3.48\pm0.59$ & $96.0\%$ \\
D3: RGB Hallucination-free
  & $98.0\%$       & ---
  & $93.3\%$       & ---      \\
D4: Specificity
  & $3.42\pm0.61$ & $99.3\%$
  & $3.35\pm0.64$ & $97.3\%$ \\
D5: Emissivity$^\dagger$
  & ---            & ---
  & $2.98\pm0.74$ & $89.3\%$ \\
 \rowcolor{blue!8} Overall Approval
  & \multicolumn{2}{c}{$\mathbf{97.3\%}$
    $[94.1, 99.0]$}
  & \multicolumn{2}{c}{$\mathbf{84.0\%}$
    $[77.8, 89.0]$} \\
\end{tabular}
\end{center}

\end{table}
\footnotetext{Pass thresholds: D1,\,D2\,$\geq$\,3;
D3\,$=$\,0; D4,\,D5\,$\geq$\,2.
Krippendorff's $\alpha{>}0.60$ on all dimensions;
$\alpha{>}0.80$ for D3.}

\subsection{Applying T-CLIP to Thermal Image Generation}
\label{sec:imggen}

CLIP is widely used in many downstream tasks, including segmentation, 
detection, and text-to-image generation models like Stable Diffusion. 
However, standard CLIP struggles with thermal domain understanding 
because it was trained exclusively on RGB images. We explore whether T-CLIP can serve as a plug-and-play replacement for 
the text encoder in SDXL \citep{SDXL}. Specifically, we replace the original CLIP 
ViT-L/14 text encoder in SDXL with our T-CLIP text encoder. Because 
SDXL's UNet was trained only on RGB images and cannot directly 
generate thermal outputs, we fine-tune the UNet on KAIST thermal 
image–caption pairs to adapt its decoding capabilities to the thermal 
domain.

\begin{figure}[h]
\begin{center}
\includegraphics[width=\linewidth]{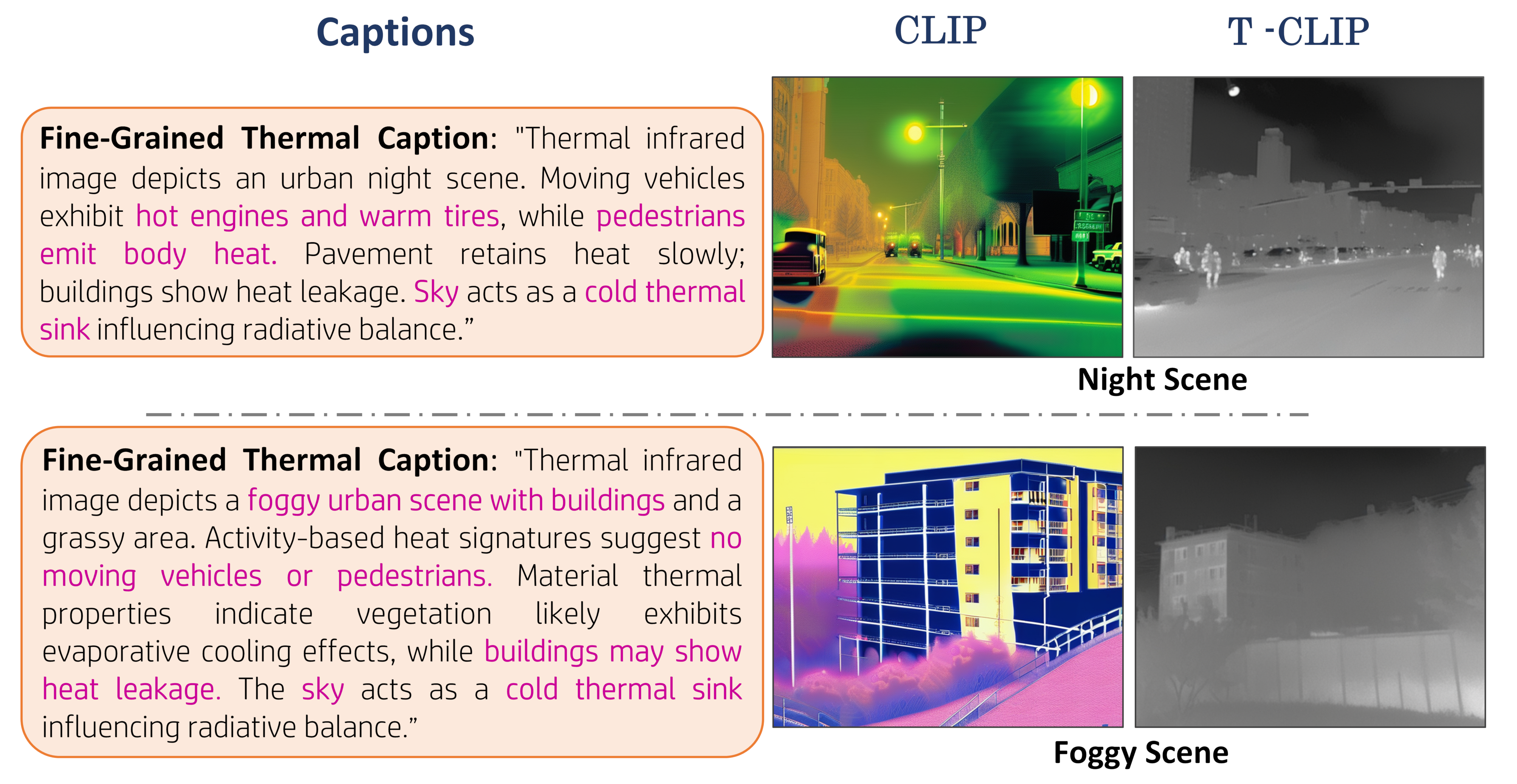}
\end{center}
\caption{T-CLIP as a plug-and-play replacement for the text encoder
in SDXL for thermal image generation. SDXL equipped with the standard
CLIP encoder produces pseudo-coloured RGB-like outputs lacking
meaningful thermal characteristics. In contrast, replacing the CLIP
encoder with T-CLIP enables SDXL to generate physically plausible
thermal images that capture pedestrian body heat, vehicle heat
emissions in nighttime scenes, and building heat leakage under foggy
conditions.}
\label{fig:special_images}
\end{figure}

Qualitatively, as shown in Figure~\ref{fig:special_images}, T-CLIP generates 
physically plausible thermal phenomena — vehicle heat emissions, 
pedestrian body heat, and atmospheric scattering, under challenging 
conditions including night scenes and foggy environments. Additional 
generated samples across different illumination and weather conditions 
are presented in section~\ref{sec:supp_generation} of the appendix.

To evaluate thermal generation quality,
we conducted a user study with
five annotators with domain expertise on 30 prompt
pairs spanning five scene categories
(full protocol and other calculations are detailed in appendix section~\ref{sec:supp_generation}). Annotators rated each generated image on Thermal
Plausibility (D1), Physics Correctness
(D2 -- a binary checklist based on thermal imaging principles, e.g.:
\emph{warm\,$\to$\,brighter; cool\,$\to$\,darker}), and Prompt
Faithfulness (D3), each on a 1--4 scale.

As shown in Table~\ref{tab:img_gen_user_study},
T-CLIP\,+\,SDXL achieves excellent ratings in terms of  Thermal Plausibility ($3.98\pm0.14$)
and substantially improved Physics
Correctness ($3.0\pm0.5$), with zero-shot
SDXL receiving a score of~1 on both D1 and D2, suggesting that
all generated images are pseudo-coloured
RGB with no visible thermal characteristics.
Prompt Faithfulness is equivalent across both models ($\Delta = +0.1$, $p > 0.05$, n.s.), suggesting that T-CLIP adds thermal quality without altering scene content generation. T-CLIP+SDXL was preferred across all 150 pairwise comparisons (100\%, $p \ll 0.001$).
Our user study demonstrates that T-CLIP's thermal representations transfer 
meaningful thermal reasoning capabilities to existing generation 
frameworks without requiring architectural changes. 
\begin{table}[t]
\caption{User study results for thermal image generation (30 pairs,
5 annotators, 150 ratings per dimension).
SDXL: zero-shot Stable Diffusion XL~\citep{podell2023sdxl} with
standard CLIP text encoder.
T-CLIP: our thermally adapted text encoder replacing the standard
CLIP encoder in SDXL.
$\Delta$ = T-CLIP score $-$ SDXL score (mean difference across 150
ratings).
$^\dagger$Sign test used because SDXL variance $=0$ (unanimous score
of 1 on D1 and D2 precludes Wilcoxon).
$^\ddagger$n.s.\ $=$ not significant ($p > 0.05$ after
Benjamini-Hochberg correction across three comparisons,
FDR $=0.05$).\protect \footnotemark}
\label{tab:img_gen_user_study}
\begin{center}
\begin{tabular}{l c c c c}
\bf DIMENSION & \bf SDXL & \bf T-CLIP & $\mathbf{\Delta}$ & \bf SIG. \\
\hline \\
D1: Thermal plausibility
  & $1.00\pm0.00$ & $3.98\pm0.14$ & $+2.98$ & $p \ll 0.001^\dagger$ \\
D2: Physics correctness
  & $1.00\pm0.00$ & $3.0\pm0.5$   & $+2.0$  & $p \ll 0.001^\dagger$ \\
D3: Prompt faithfulness
  & $2.9\pm0.4$   & $3.0\pm0.5$   & $+0.1$  & n.s.$^\ddagger$ \\
\rowcolor{blue!8} Preference
  & \multicolumn{4}{c}{100\% T-CLIP preferred
    ($n = 150$, $p \ll 0.001$)} \\
\end{tabular}
\end{center}
\end{table}
\footnotetext{Pass thresholds and statistical notes:
$^\dagger$Sign test used because SDXL variance $=0$.
$^\ddagger$n.s.\ $=$ not significant ($p > 0.05$,
Benjamini-Hochberg corrected, FDR $=0.05$).}
\section{Conclusion, Limitations, and Future Work}

We introduced IR-Cap, the first physics-aware thermal captioning
pipeline and dataset providing complementary global and fine-grained
thermal descriptions across three public benchmarks, and T-CLIP, a
decoupled dual-LoRA framework for thermal CLIP adaptation.
The central insight of this work is that global scene-level thermal
context and object-level heat signatures are geometrically
incompatible and cannot be meaningfully optimized within a shared
embedding space.
This finding, validated through feature geometry analysis and controlled ablations, directly motivates our decoupled design: two independent specialized branches trained separately on IR-Cap's global and fine-grained captions, then fused at inference.
T-CLIP achieves consistent improvements across three thermal
benchmarks on cross-modal retrieval and offers preliminary evidence of its transferability to text-conditioned thermal image generation. We release IR-Cap and our captioning pipeline to facilitate future
progress in this area.

\paragraph{Limitations and Future Work.}
IR-Cap relies on paired RGB images as
semantic anchors for generating thermal
captions, which limits the modelling of
fine-grained thermal phenomena not directly
observable in the visible spectrum.
This motivates future work on richer thermal
metadata annotation strategies to enable
more precise characterization of object-level heat signatures. While T-CLIP can be used for text-conditioned
thermal image generation as a plug-and-play
text encoder replacement, a systematic
study of thermal image generation constitutes a substantial independent
research direction beyond the scope of
this work.


\bibliography{main}
\bibliographystyle{tmlr}

\appendix
\section{Appendix}


\subsection{Quantifying the Thermal Perception Gap}
\label{sec:supp_cosine}
 
Table~\ref{tab:supp_welch_bh} reports the full pairwise Welch's
$t$-test results for the image-text cosine similarity distributions
shown in sections~\ref{sec:intro} and~\ref{sec:challenges}, including
BH-corrected $p$-values. All three pairwise differences are
statistically significant with $p \ll 0.001$ before and after
correction, confirming the robustness of the reported cosine
similarity comparisons.

\begin{table}[t]
\caption{Pairwise Welch's $t$-test results for Figure~1 (main
manuscript) image-text cosine similarity distributions on the
KAIST~\citep{hwang2015multispectral} test set.
Benjamini-Hochberg (BH) correction applied across all three
comparisons (FDR\,$=$\,0.05).
$n{=}2{,}252$ image-caption pairs per method.
$p$-values are unchanged in practical terms after correction,
confirming the negligible effect of adjustment given effect sizes
of this magnitude.\protect \footnotemark}
\label{tab:supp_welch_bh}
\begin{center}
\resizebox{\linewidth}{!}{%
\begin{tabular}{l c c c c c c}
\bf COMPARISON
  & \bf MEAN A & \bf MEAN B
  & $\mathbf{\Delta}$ & $\mathbf{t}$\textbf{-STAT}
  & $p_{\text{RAW}}$ & $p_{\text{BH}}$ \\
\hline \\
Zero-shot CLIP vs Global LoRA
  & 0.3449 & 0.3322 & $-0.0127$ & $11.559$
  & $2.83\times10^{-30}$  & $2.83\times10^{-30}$  \\
Global LoRA vs T-CLIP
  & 0.3322 & 0.3716 & $+0.0394$ & $-30.350$
  & $1.86\times10^{-183}$ & $5.58\times10^{-183}$ \\
Zero-shot CLIP vs T-CLIP
  & 0.3449 & 0.3716 & $+0.0267$ & $-29.132$
  & $7.61\times10^{-167}$ & $1.14\times10^{-166}$ \\
\end{tabular}
}
\end{center}
\end{table}
\footnotetext{Method means (cosine similarity):
Zero-shot CLIP\,$=$\,$0.3449{\pm}0.0203$;
Global LoRA\,$=$\,$0.3322{\pm}0.0482$;
T-CLIP\,$=$\,$0.3716{\pm}0.0384$.
$\Delta{=}$\,Mean~B\,$-$\,Mean~A.
Two-sided Welch's $t$-test (unequal variance).}

\subsection{IR-Cap Dataset --- Instruction Prompts}
\label{sec:supp_dataset}

\begin{figure}[h]
\begin{center}

\includegraphics[width=\linewidth]{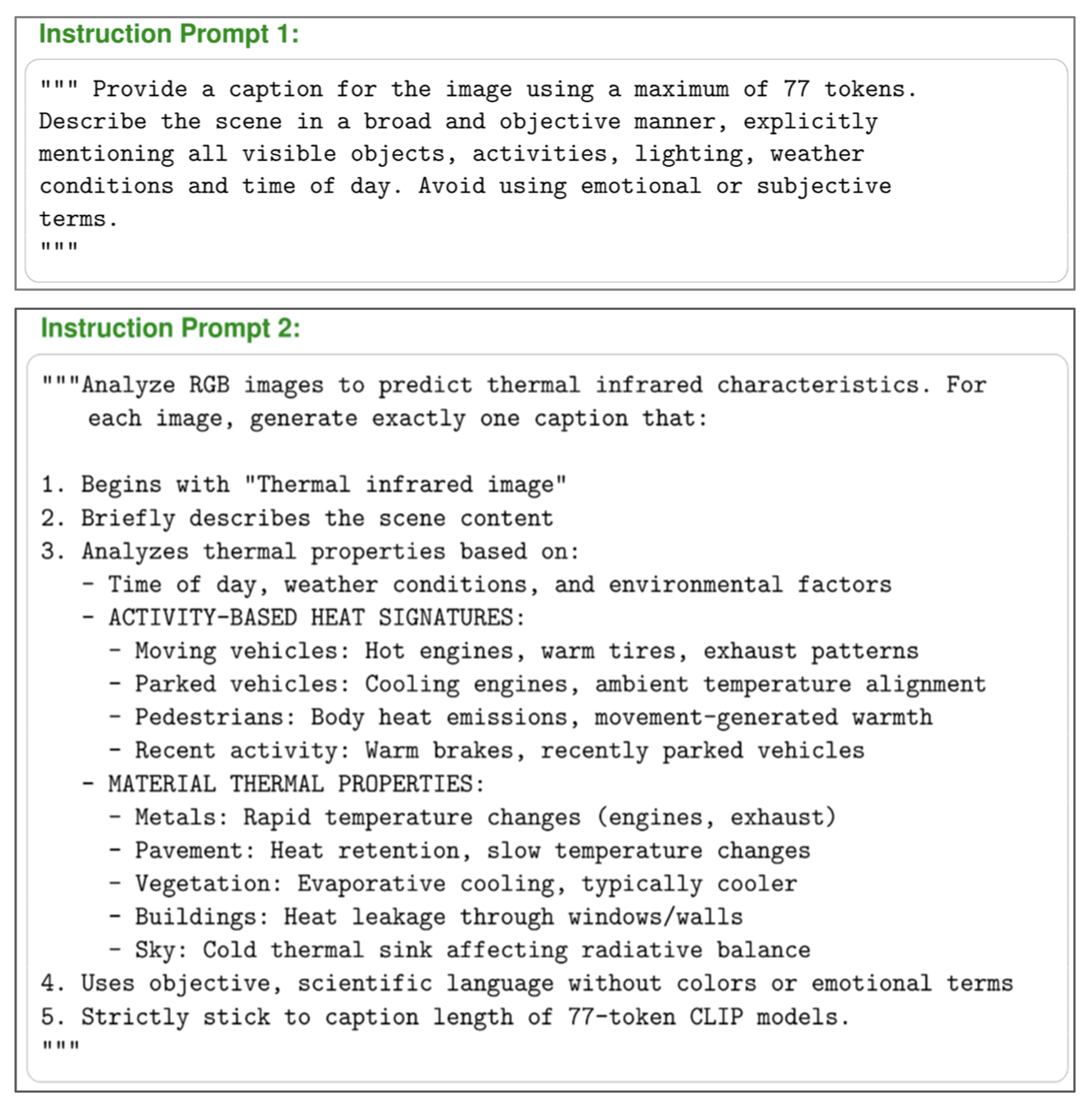}
\caption{Instruction prompts for IR-Cap caption generation pipeline.
         Instruction Prompt~1 generates Global Thermal Captions
         describing scene-level environmental context.
         Instruction Prompt~2 generates Fine-Grained Thermal Captions
         encoding object-level heat signatures and thermal phenomena.}
\label{fig:suppl_prompts}
\end{center}
\end{figure}

As discussed in section~\ref{sec:ircap} we employed a dual prompting strategy with Qwen2.5-VL-72B-Instruct,
using the visible-spectrum images as semantic context for
generating captions for corresponding thermal images.
Figure~\ref{fig:suppl_prompts} shows the two instruction prompts used for
Global and Fine-Grained caption generation respectively.

\subsection{IR-Cap Human Evaluation}
\label{sec:supp_ircap}

This section provides complete protocol details for the human evaluation of the IR-Cap dataset, supporting section~\ref{sec:ircap_eval}. It includes the full rubric, stratified sampling design, annotator assignment matrix, failure case analysis, and inter-annotator agreement.

\subsubsection{Full Rubric Anchor Descriptions}
\label{sec:supp_rubric}

Table~\ref{tab:supp_rubric} provide complete anchor
descriptions and examples for all rating dimensions.
Annotators were provided this table as a printed reference sheet.
D3 is binary (0/1); D5 applies to Fine-Grained captions only.
Pass thresholds: D1,\,D2\,$\geq$\,3; D3\,$=$\,0; D4,\,D5\,$\geq$\,2.


\begin{table}[t]
\caption{Full anchor descriptions for D1 (thermal accuracy),
D2 (semantic correctness), D3 (RGB hallucination),
D4 (specificity), and D5 (emissivity reasoning).
Pass thresholds: D1, D2 $\geq$ 3; D3 $=$ 0; D4, D5 $\geq$ 2.
D5 applies to fine-grained captions only.
Overall approval requires passing all applicable
dimensions simultaneously.}
\label{tab:supp_rubric}
\begin{center}
\resizebox{\linewidth}{!}{%
\begin{tabular}{c p{2.5cm} p{4.0cm} p{7.0cm}}
\bf DIM. & \bf SCORE & \bf DESCRIPTION & \bf EXAMPLE \\
\hline \\
\rowcolor{black!6}\multicolumn{4}{l}{\textit{D1: Thermal accuracy (1--4, pass $\geq$ 3)}} \\
\multirow[t]{4}{*}{D1}
  & 1 & Physically impossible thermal claims
    & ``The sky radiates heat downward, warming the wet pavement'' \\[2pt]
  & 2 & Thermally neutral; no physics reasoning
    & ``People and cars are visible on a road at night'' \\[2pt]
  & 3 & Correct thermal reasoning; incomplete
    & ``Scene shows warm objects against cool background at night'' \\[2pt]
  & 4 & Full correct thermal reasoning
    & ``Night scene: reduced ambient temperature; wet pavement retains
       residual heat; overcast sky acts as cold thermal sink'' \\
\rowcolor{black!6}\multicolumn{4}{l}{\textit{D2: Semantic correctness (1--4, pass $\geq$ 3)}} \\
\multirow[t]{4}{*}{D2}
  & 1 & Wrong scene described entirely
    & Caption describes indoor scene; image is outdoor road \\[2pt]
  & 2 & Correct environment; wrong objects
    & ``Cars on highway'' but image shows pedestrians \\[2pt]
  & 3 & Correct objects; minor attribute errors
    & ``Two pedestrians'' but image shows three \\[2pt]
  & 4 & All objects and scene correctly identified
    & --- \\
\rowcolor{black!6}\multicolumn{4}{l}{\textit{D3: RGB hallucination (binary 0/1, pass $=$ 0)}} \\
\multirow[t]{2}{*}{D3}
  & 0 & No RGB-style colour or texture language used
    & --- \\[2pt]
  & 1 & RGB hallucination detected
    & ``White lane markings''; ``Yellow streetlights'';
      ``Dark coloured car'' \\
\rowcolor{black!6}\multicolumn{4}{l}{\textit{D4: Specificity (1--4, pass $\geq$ 2)}} \\
\multirow[t]{4}{*}{D4}
  & 1 & Completely generic
    & ``An infrared thermal image'' \\[2pt]
  & 2 & Scene identified; no thermal detail
    & ``People walking on a road at night'' \\[2pt]
  & 3 & Some thermal context present
    & ``Night scene with warm pedestrians and cooler background'' \\[2pt]
  & 4 & Rich environmental thermal detail
    & ``Late evening urban scene; ambient temperature low; wet road
       retains residual heat; buildings show heat leakage'' \\
\rowcolor{black!6}\multicolumn{4}{l}{\textit{D5: Emissivity reasoning (1--4, pass $\geq$ 2) --- fine-grained only}} \\
\multirow[t]{4}{*}{D5}
  & 1 & No object-level thermal reasoning
    & ``Thermal image of people and cars'' \\[2pt]
  & 2 & Generic warm/cool labels only
    & ``People are warm, cars are cool'' \\[2pt]
  & 3 & Material-specific reasoning; incomplete
    & ``Moving vehicles show warm engines; pedestrians emit body
       heat from torso'' \\[2pt]
  & 4 & Full emissivity and physics reasoning
    & ``Moving vehicles exhibit hot engine blocks and warm tires
       from friction; pedestrians emit body heat at torso;
       vegetation appears cool due to evaporative cooling'' \\
\end{tabular}
}
\end{center}
\end{table}

\paragraph{Approval rate computation.}
The per-dimension approval rate (Appr.\%) for dimension $D_i$ is:
\begin{equation}
  \text{Appr.}_{D_i} =
  \frac{|\{c : s_{D_i}(c) \geq \tau_i\}|}{n},
  \label{eq:appr_dim}
\end{equation}
where $s_{D_i}(c)$ is the mean annotator score for caption $c$,
$\tau_i$ is the pass threshold, and $n{=}150$.
Overall Approval requires passing \emph{all applicable} dimensions
simultaneously:
\begin{equation}
  \text{Overall Appr.} =
  \frac{|\{c : \forall\,i \in \mathcal{D}(c),\;
             s_{D_i}(c) \geq \tau_i\}|}{n},
  \label{eq:appr_overall}
\end{equation}
where $\mathcal{D}(c)$ denotes the applicable dimensions for caption
$c$ (D1--D4 for Global; D1--D5 for Fine-Grained).
Overall Approval is at most the minimum per-dimension approval rate
and is not the average of per-dimension rates.

\subsubsection{Stratified Sampling Design}
\label{sec:supp_sampling}

Samples were drawn uniformly across dataset origin, time of day,
and scene type; Table~\ref{tab:supp_sampling} details the breakdown.
\begin{table}[t]
\caption{Stratified sampling design across
KAIST~\citep{hwang2015multispectral},
FLIR~\citep{flir}, and
FMB~\citep{liu2023multi},
time of day, and scene type.}
\label{tab:supp_sampling}
\begin{center}
\begin{tabular}{l l c c c c}
\bf CAPTION TYPE & \bf DATASET
  & \bf DAY & \bf NIGHT & \bf OTHER & \bf TOTAL \\
\hline \\
\multirow{4}{*}{Global (150)}
  & KAIST & 17 & 17 & 16 & 50 \\
  & FLIR  & 17 & 17 & 16 & 50 \\
  & FMB   & 17 & 17 & 16 & 50 \\
  \rowcolor{blue!6}&Total & 51 & 51 & 48 & 150 \\ \\ 
\multirow{4}{*}{Fine-grained (150)}
  & KAIST & 17 & 17 & 16 & 50 \\
  & FLIR  & 17 & 17 & 16 & 50 \\
  & FMB   & 17 & 17 & 16 & 50 \\
 \rowcolor{blue!6} & Total & 51 & 51 & 48 & 150 \\ \\
\rowcolor{blue!6}\multicolumn{2}{l}{Grand total}
  &102 & 102 & 96 & 300 \\
\end{tabular}
\end{center}
\end{table}

\subsubsection{Annotator Assignment Matrix}
\label{sec:supp_assignment}

Each annotator evaluates 63 items per caption type (126 total),
comprising 20 shared overlap items and approximately 43 unique items
per batch, yielding an average of 3.3 ratings per item.
For items where annotator scores differed by more than one point on
any dimension, a sixth expert adjudicator provided a tie-breaking
score (7 items, 2.3\% of overlap items).

\begin{table}[t]
\caption{Annotator assignment matrix, applied independently to both
global and fine-grained sets.
\ding{51}\,=\,assigned; ---\,=\,not assigned.}
\label{tab:supp_assignment}
\begin{center}
\begin{tabular}{l c c c c c c}
\bf ITEM SET
  & \bf ANN.~1 & \bf ANN.~2 & \bf ANN.~3
  & \bf ANN.~4 & \bf ANN.~5 & \bf ITEMS \\
\hline \\
Overlap (items 1--20)
  & \ding{51} & \ding{51} & \ding{51} & \ding{51} & \ding{51} & 20 \\
Batch A (items 21--63)
  & \ding{51} & \ding{51} & \ding{51} & ---        & ---        & 43 \\
Batch B (items 64--107)
  & ---        & \ding{51} & \ding{51} & \ding{51} & ---        & 44 \\
Batch C (items 108--150)
  & ---        & ---        & \ding{51} & \ding{51} & \ding{51} & 43 \\
\rowcolor{blue!6}Per annotator & 63 & 63 & 63 & 63 & 63 & \\
\end{tabular}
\end{center}
\end{table}

\subsubsection{Failure Case Analysis}
\label{sec:supp_failures}

Table~\ref{tab:supp_failures} summarises the three systematic failure
categories identified in the 24 Fine-Grained captions failing
overall approval, all arising from thermal phenomena not directly
observable in the RGB semantic anchor.
\begin{table}[t]
\caption{Representative limitation cases in fine-grained IR-Cap
captions arising from the RGB-anchoring strategy.
Percentage of failures (24 total) per category
shown in parentheses.}
\label{tab:supp_failures}
\begin{center}
\begin{tabular}{p{2.5cm} p{3.8cm} p{3.8cm} p{2.4cm}}
\bf CATEGORY (SHARE) & \bf GENERATED CAPTION
  & \bf IDEAL CAPTION & \bf ROOT CAUSE \\
\hline \\
State ambiguity (54\%)
  & ``Vehicles exhibit warm engines and hot tires from friction''
  & ``Moving vehicles exhibit hot engine blocks; stationary vehicles
    show cooling engines with reduced heat signatures''
  & Motion state not inferable from RGB; VLMs cannot map motion
    to thermal signatures \\
Intra-object distribution (29\%)
  & ``Pedestrians emit body heat''
  & ``Torso and head show strong heat emission; extremities cooler;
    clothing acts as partial thermal insulation''
  & Within-object thermal gradients not inferable in RGB \\
Emissivity mechanism (17\%)
  & ``Wet road retains heat from daytime solar radiation''
  & ``Wet surface shows elevated emissivity (${\approx}0.97$ vs.\
    dry ${\approx}0.92$), appearing thermally brighter independent
    of actual temperature''
  & Emissivity physics concept absent from VLM training distribution \\
\end{tabular}
\end{center}
\end{table}

\subsubsection{Pairwise Annotator Agreement}
\label{sec:supp_agreement}

All five annotators achieved substantial pairwise agreement
(Krippendorff's $\alpha{>}0.60$) across all dimensions on the
20 overlap items, consistent with the overall $\alpha$ values
reported in the main paper.
D3 (RGB Hallucination) consistently achieved the highest pairwise
agreement ($\alpha{>}0.80$), reflecting both its binary scale and
the straightforward nature of the judgment: annotators simply check
whether the caption contains explicit colour or texture terms
(e.g., ``white'', ``yellow'', ``dark'') that are meaningless in
the thermal infrared domain.

\paragraph{Data Availability.}
All caption pairs and images used for the human evaluation of the IR-Cap dataset are available at
\url{https://drive.google.com/drive/folders/10Xt5lVwSrkiSvJdmFRFMBZtg0SRUvhhS?usp=drive_link}.

\subsection{Retrieval Results with Multi-Seed Statistics}
\label{sec:supp_retrieval}

Tables~\ref{tab:supp_kaist_full},~\ref{tab:supp_flir_full} and~\ref{tab:supp_fmb_full}
report the full Recall@K retrieval results across all three benchmarks.
R@K is reported as mean\,$\pm$\,std across three independent training
runs (seeds 0, 42, 123).
Zero-shot CLIP results are deterministic and reported
as point estimates only.



\begin{table}[t]
\caption{Text-image retrieval recall@K on the KAIST~\citep{hwang2015multispectral}
  test set. All values reported as mean\,$\pm$\,std across three independent
  training runs. \textbf{Bold}: best result. \underline{Underline}: second best.
  F-G = fine-grained. $\Delta$: relative R@K gain of T-CLIP (Dual) over
  the strongest baseline (Global LoRA).}
\label{tab:supp_kaist_full}
\begin{center}
\resizebox{\linewidth}{!}{%
\begin{tabular}{l c c c c c c}
\rowcolor{black!10}
\multicolumn{7}{c}{\bf IMAGE-TO-TEXT RETRIEVAL} \\ 
\multicolumn{1}{c}{\bf METHOD}
  & \textbf{R@1} & \textbf{R@5} & \textbf{R@10}
  & \textbf{R@25} & \textbf{R@50} & \textbf{R@100} \\
\hline \\
CLIP (Global)
  & 0.003 & 0.016 & 0.034 & 0.068 & 0.110 & 0.179 \\
CLIP (F-G)
  & 0.001 & 0.004 & 0.008 & 0.029 & 0.057 & 0.096 \\
CLIP-Adapter (Global)
  & $0.018{\pm}0.001$ & $0.072{\pm}0.004$ & $0.123{\pm}0.002$
  & $0.229{\pm}0.003$ & $0.357{\pm}0.003$ & $0.521{\pm}0.002$ \\
CLIP-Adapter (F-G)
  & $0.009{\pm}0.002$ & $0.041{\pm}0.002$ & $0.081{\pm}0.005$
  & $0.167{\pm}0.006$ & $0.283{\pm}0.005$ & $0.436{\pm}0.006$ \\
DeCLIP (Global)
  & $0.038{\pm}0.002$ & $0.129{\pm}0.001$ & $0.206{\pm}0.009$
  & $0.349{\pm}0.005$ & $0.497{\pm}0.008$ & $0.644{\pm}0.004$ \\
DeCLIP (F-G)
  & $0.013{\pm}0.000$ & $0.048{\pm}0.002$ & $0.088{\pm}0.002$
  & $0.169{\pm}0.002$ & $0.273{\pm}0.008$ & $0.433{\pm}0.013$ \\
Global LoRA
  & $0.070{\pm}0.002$ & $0.224{\pm}0.004$ & $0.332{\pm}0.013$
  & $0.520{\pm}0.007$ & $0.685{\pm}0.015$ & $0.825{\pm}0.013$ \\
F-G LoRA
  & $0.021{\pm}0.002$ & $0.073{\pm}0.007$ & $0.125{\pm}0.005$
  & $0.248{\pm}0.010$ & $0.381{\pm}0.006$ & $0.571{\pm}0.003$ \\
T-CLIP (Global)
  & $\underline{0.071{\pm}0.003}$ & $\underline{0.233{\pm}0.002}$
  & $\underline{0.349{\pm}0.003}$ & $\underline{0.541{\pm}0.004}$
  & $\underline{0.695{\pm}0.006}$ & $\underline{0.837{\pm}0.011}$ \\
T-CLIP (F-G)
  & $0.016{\pm}0.002$ & $0.064{\pm}0.003$ & $0.101{\pm}0.007$
  & $0.194{\pm}0.007$ & $0.311{\pm}0.010$ & $0.473{\pm}0.015$ \\
\textbf{T-CLIP (Dual)}
  & $\mathbf{0.078{\pm}0.004}$ & $\mathbf{0.252{\pm}0.003}$
  & $\mathbf{0.374{\pm}0.004}$ & $\mathbf{0.571{\pm}0.008}$
  & $\mathbf{0.729{\pm}0.013}$ & $\mathbf{0.862{\pm}0.009}$ \\ 
\rowcolor{blue!6}
$\Delta$ vs.\ Global LoRA
  & +10.5\% & +12.4\% & +12.7\% & +9.7\% & +6.3\% & +4.5\% \\ \\
\rowcolor{black!10}
\multicolumn{7}{c}{\bf TEXT-TO-IMAGE RETRIEVAL} \\
\multicolumn{1}{c}{\bf METHOD}
  & \textbf{R@1} & \textbf{R@5} & \textbf{R@10}
  & \textbf{R@25} & \textbf{R@50} & \textbf{R@100} \\
\hline \\
CLIP (Global)
  & 0.002 & 0.013 & 0.027 & 0.048 & 0.080 & 0.135 \\
CLIP (F-G)
  & 0.001 & 0.007 & 0.014 & 0.030 & 0.048 & 0.088 \\
CLIP-Adapter (Global)
  & $0.020{\pm}0.002$ & $0.070{\pm}0.002$ & $0.119{\pm}0.003$
  & $0.234{\pm}0.006$ & $0.360{\pm}0.001$ & $0.538{\pm}0.003$ \\
CLIP-Adapter (F-G)
  & $0.010{\pm}0.000$ & $0.046{\pm}0.000$ & $0.082{\pm}0.003$
  & $0.166{\pm}0.004$ & $0.281{\pm}0.008$ & $0.445{\pm}0.004$ \\
DeCLIP (Global)
  & $0.041{\pm}0.000$ & $0.133{\pm}0.004$ & $0.209{\pm}0.002$
  & $0.357{\pm}0.005$ & $0.496{\pm}0.014$ & $0.650{\pm}0.005$ \\
DeCLIP (F-G)
  & $0.014{\pm}0.000$ & $0.056{\pm}0.002$ & $0.094{\pm}0.001$
  & $0.175{\pm}0.006$ & $0.285{\pm}0.013$ & $0.439{\pm}0.010$ \\
Global LoRA
  & $0.069{\pm}0.005$ & $0.221{\pm}0.008$ & $0.331{\pm}0.010$
  & $0.521{\pm}0.013$ & $0.679{\pm}0.012$ & $0.822{\pm}0.010$ \\
F-G LoRA
  & $0.019{\pm}0.002$ & $0.079{\pm}0.010$ & $0.139{\pm}0.008$
  & $0.260{\pm}0.002$ & $0.400{\pm}0.007$ & $0.582{\pm}0.012$ \\
T-CLIP (Global)
  & $\underline{0.078{\pm}0.002}$ & $\underline{0.242{\pm}0.004}$
  & $\underline{0.359{\pm}0.009}$ & $\underline{0.554{\pm}0.008}$
  & $\underline{0.710{\pm}0.010}$ & $\underline{0.843{\pm}0.009}$ \\
T-CLIP (F-G)
  & $0.012{\pm}0.002$ & $0.048{\pm}0.004$ & $0.084{\pm}0.006$
  & $0.157{\pm}0.010$ & $0.255{\pm}0.012$ & $0.394{\pm}0.018$ \\
\textbf{T-CLIP (Dual)}
  & $\mathbf{0.084{\pm}0.002}$ & $\mathbf{0.264{\pm}0.005}$
  & $\mathbf{0.386{\pm}0.010}$ & $\mathbf{0.580{\pm}0.008}$
  & $\mathbf{0.740{\pm}0.007}$ & $\mathbf{0.870{\pm}0.011}$ \\ 
\rowcolor{blue!6}
$\Delta$ vs.\ Global LoRA
  & +21.9\% & +19.4\% & +16.8\% & +11.3\% & +9.1\% & +5.8\% \\
\end{tabular}
}
\end{center}
\end{table}

\begin{table}[t]
\caption{Text-image retrieval recall@K on the FLIR~\citep{flir}
  test set. All values reported as mean\,$\pm$\,std across three independent
  training runs. \textbf{Bold}: best result. \underline{Underline}: second best.
  F-G = fine-grained. $\Delta$: relative R@K gain of T-CLIP (Dual) over
  the strongest baseline (Global LoRA).}
\label{tab:supp_flir_full}
\begin{center}
\resizebox{\linewidth}{!}{%
\begin{tabular}{l c c c c c c}
\rowcolor{black!10}
\multicolumn{7}{c}{\bf IMAGE-TO-TEXT RETRIEVAL} \\
\multicolumn{1}{c}{\bf METHOD}
  & \textbf{R@1} & \textbf{R@5} & \textbf{R@10}
  & \textbf{R@25} & \textbf{R@50} & \textbf{R@100} \\
\hline \\
CLIP (Global)
  & 0.014 & 0.070 & 0.115 & 0.242 & 0.360 & 0.513 \\
CLIP (F-G)
  & 0.006 & 0.022 & 0.039 & 0.110 & 0.220 & 0.332 \\
CLIP-Adapter (Global)
  & $0.045{\pm}0.006$ & $0.153{\pm}0.005$ & $0.247{\pm}0.010$
  & $0.411{\pm}0.016$ & $0.545{\pm}0.014$ & $0.700{\pm}0.014$ \\
CLIP-Adapter (F-G)
  & $0.027{\pm}0.003$ & $0.081{\pm}0.001$ & $0.134{\pm}0.003$
  & $0.250{\pm}0.005$ & $0.402{\pm}0.012$ & $0.582{\pm}0.005$ \\
DeCLIP (Global)
  & $0.049{\pm}0.002$ & $0.154{\pm}0.002$ & $0.228{\pm}0.005$
  & $0.376{\pm}0.005$ & $0.507{\pm}0.002$ & $0.642{\pm}0.004$ \\
DeCLIP (F-G)
  & $0.012{\pm}0.001$ & $0.056{\pm}0.003$ & $0.104{\pm}0.008$
  & $0.197{\pm}0.002$ & $0.312{\pm}0.001$ & $0.474{\pm}0.010$ \\
Global LoRA
  & $0.105{\pm}0.003$ & $0.329{\pm}0.011$ & $0.487{\pm}0.008$
  & $0.678{\pm}0.003$ & $0.816{\pm}0.004$ & $0.914{\pm}0.002$ \\
F-G LoRA
  & $0.027{\pm}0.004$ & $0.089{\pm}0.007$ & $0.150{\pm}0.008$
  & $0.290{\pm}0.016$ & $0.453{\pm}0.016$ & $0.642{\pm}0.014$ \\
T-CLIP (Global)
  & $\underline{0.119{\pm}0.004}$ & $\underline{0.353{\pm}0.007}$
  & $\underline{0.504{\pm}0.013}$ & $\underline{0.707{\pm}0.007}$
  & $\underline{0.833{\pm}0.007}$ & $\underline{0.923{\pm}0.003}$ \\
T-CLIP (F-G)
  & $0.025{\pm}0.005$ & $0.107{\pm}0.006$ & $0.182{\pm}0.004$
  & $0.316{\pm}0.009$ & $0.473{\pm}0.032$ & $0.649{\pm}0.014$ \\
\textbf{T-CLIP (Dual)}
  & $\mathbf{0.123{\pm}0.010}$ & $\mathbf{0.357{\pm}0.006}$
  & $\mathbf{0.517{\pm}0.015}$ & $\mathbf{0.727{\pm}0.005}$
  & $\mathbf{0.851{\pm}0.005}$ & $\mathbf{0.931{\pm}0.008}$ \\
\rowcolor{blue!6}
$\Delta$ vs.\ Global LoRA
  & $+$17.1\% & $+$8.5\% & $+$6.2\%
  & $+$7.2\% & $+$4.3\% & $+$1.9\% \\ \\
\rowcolor{black!10}
\multicolumn{7}{c}{\bf TEXT-TO-IMAGE RETRIEVAL} \\
\multicolumn{1}{c}{\bf METHOD}
  & \textbf{R@1} & \textbf{R@5} & \textbf{R@10}
  & \textbf{R@25} & \textbf{R@50} & \textbf{R@100} \\
\hline \\
CLIP (Global)
  & 0.014 & 0.062 & 0.106 & 0.229 & 0.319 & 0.467 \\
CLIP (F-G)
  & 0.005 & 0.019 & 0.032 & 0.093 & 0.158 & 0.253 \\
CLIP-Adapter (Global)
  & $0.031{\pm}0.003$ & $0.130{\pm}0.007$ & $0.217{\pm}0.009$
  & $0.381{\pm}0.024$ & $0.529{\pm}0.024$ & $0.706{\pm}0.017$ \\
CLIP-Adapter (F-G)
  & $0.019{\pm}0.001$ & $0.075{\pm}0.006$ & $0.133{\pm}0.011$
  & $0.244{\pm}0.005$ & $0.367{\pm}0.009$ & $0.555{\pm}0.009$ \\
DeCLIP (Global)
  & $0.045{\pm}0.002$ & $0.158{\pm}0.011$ & $0.232{\pm}0.001$
  & $0.367{\pm}0.001$ & $0.510{\pm}0.008$ & $0.641{\pm}0.005$ \\
DeCLIP (F-G)
  & $0.015{\pm}0.001$ & $0.073{\pm}0.003$ & $0.114{\pm}0.008$
  & $0.218{\pm}0.012$ & $0.324{\pm}0.010$ & $0.474{\pm}0.001$ \\
Global LoRA
  & $0.106{\pm}0.003$ & $0.321{\pm}0.007$ & $0.459{\pm}0.006$
  & $0.673{\pm}0.006$ & $0.812{\pm}0.007$ & $0.907{\pm}0.006$ \\
F-G LoRA
  & $0.035{\pm}0.003$ & $0.127{\pm}0.005$ & $0.206{\pm}0.003$
  & $0.369{\pm}0.013$ & $0.528{\pm}0.024$ & $0.702{\pm}0.015$ \\
T-CLIP (Global)
  & $\underline{0.105{\pm}0.007}$ & $\underline{0.348{\pm}0.002}$
  & $\underline{0.492{\pm}0.005}$ & $\underline{0.712{\pm}0.001}$
  & $\underline{0.839{\pm}0.008}$ & $\underline{0.925{\pm}0.004}$ \\
T-CLIP (F-G)
  & $0.020{\pm}0.004$ & $0.069{\pm}0.004$ & $0.130{\pm}0.009$
  & $0.239{\pm}0.012$ & $0.361{\pm}0.013$ & $0.523{\pm}0.025$ \\
\textbf{T-CLIP (Dual)}
  & $\mathbf{0.117{\pm}0.003}$ & $\mathbf{0.364{\pm}0.008}$
  & $\mathbf{0.511{\pm}0.005}$ & $\mathbf{0.728{\pm}0.005}$
  & $\mathbf{0.851{\pm}0.006}$ & $\mathbf{0.932{\pm}0.005}$ \\
\rowcolor{blue!6}
$\Delta$ vs.\ Global LoRA
  & $+$10.4\% & $+$13.4\% & $+$11.3\%
  & $+$8.2\% & $+$4.8\% & $+$2.8\% \\
\end{tabular}
}
\end{center}
\end{table}


\begin{table}[t]
\caption{Text-image retrieval recall@K on the FMB~\citep{liu2023multi}
  test set. All values reported as mean\,$\pm$\,std across three independent
  training runs. \textbf{Bold}: best result. \underline{Underline}: second
  best. F-G = fine-grained. $\Delta$: relative R@K gain of the best
  T-CLIP variant per metric over Global LoRA.}
\label{tab:supp_fmb_full}
\begin{center}
\resizebox{\linewidth}{!}{%
\begin{tabular}{l c c c c c c}
\rowcolor{black!10}
\multicolumn{7}{c}{\bf IMAGE-TO-TEXT RETRIEVAL} \\
\multicolumn{1}{c}{\bf METHOD}
  & \textbf{R@1} & \textbf{R@5} & \textbf{R@10}
  & \textbf{R@25} & \textbf{R@50} & \textbf{R@100} \\
\hline \\
CLIP (Global)
  & 0.043 & 0.118 & 0.175 & 0.318 & 0.443 & 0.632 \\
CLIP (F-G)
  & 0.011 & 0.082 & 0.143 & 0.225 & 0.279 & 0.532 \\
CLIP-Adapter (Global)
  & $0.075{\pm}0.007$ & $0.266{\pm}0.005$ & $0.409{\pm}0.009$
  & $0.605{\pm}0.002$ & $0.768{\pm}0.000$ & $0.905{\pm}0.002$ \\
CLIP-Adapter (F-G)
  & $0.029{\pm}0.004$ & $0.157{\pm}0.011$ & $0.277{\pm}0.005$
  & $0.455{\pm}0.016$ & $0.677{\pm}0.002$ & $0.880{\pm}0.009$ \\
DeCLIP (Global)
  & $0.095{\pm}0.013$ & $0.268{\pm}0.004$ & $0.400{\pm}0.007$
  & $0.573{\pm}0.002$ & $0.704{\pm}0.004$ & $0.834{\pm}0.009$ \\
DeCLIP (F-G)
  & $0.041{\pm}0.009$ & $0.145{\pm}0.009$ & $0.238{\pm}0.002$
  & $0.402{\pm}0.005$ & $0.561{\pm}0.007$ & $0.743{\pm}0.018$ \\
Global LoRA
  & $\underline{0.104{\pm}0.015}$ & $\underline{0.342{\pm}0.012}$ & $0.496{\pm}0.016$
  & $0.731{\pm}0.019$ & $0.866{\pm}0.012$ & $0.956{\pm}0.010$ \\
F-G LoRA
  & $0.037{\pm}0.002$ & $0.133{\pm}0.014$ & $0.248{\pm}0.015$
  & $0.479{\pm}0.038$ & $0.658{\pm}0.032$ & $0.842{\pm}0.015$ \\
T-CLIP (Global)
  & $\mathbf{0.105{\pm}0.012}$ & $\mathbf{0.362{\pm}0.007}$ & $\mathbf{0.539{\pm}0.018}$
  & $\underline{0.762{\pm}0.005}$ & $\underline{0.871{\pm}0.013}$ & $\underline{0.958{\pm}0.002}$ \\
T-CLIP (F-G)
  & $0.042{\pm}0.002$ & $0.151{\pm}0.014$ & $0.245{\pm}0.002$
  & $0.413{\pm}0.009$ & $0.580{\pm}0.016$ & $0.788{\pm}0.012$ \\
\textbf{T-CLIP (Dual)}
  & $0.098{\pm}0.017$ & $0.342{\pm}0.012$ & $\underline{0.537{\pm}0.014}$
  & $\mathbf{0.771{\pm}0.008}$ & $\mathbf{0.895{\pm}0.007}$ & $\mathbf{0.973{\pm}0.005}$ \\
\rowcolor{blue!6}
$\Delta$ vs.\ Global LoRA
  & $+$1.0\% & $+$5.8\% & $+$8.7\%
  & $+$5.5\% & $+$3.3\% & $+$1.8\% \\ \\
\rowcolor{black!10}
\multicolumn{7}{c}{\bf TEXT-TO-IMAGE RETRIEVAL} \\
\multicolumn{1}{c}{\bf METHOD}
  & \textbf{R@1} & \textbf{R@5} & \textbf{R@10}
  & \textbf{R@25} & \textbf{R@50} & \textbf{R@100} \\
\hline \\
CLIP (Global)
  & 0.018 & 0.071 & 0.143 & 0.243 & 0.418 & 0.618 \\
CLIP (F-G)
  & 0.004 & 0.032 & 0.068 & 0.157 & 0.289 & 0.479 \\
CLIP-Adapter (Global)
  & $0.063{\pm}0.002$ & $0.259{\pm}0.009$ & $0.402{\pm}0.009$
  & $0.613{\pm}0.009$ & $0.775{\pm}0.018$ & $0.927{\pm}0.013$ \\
CLIP-Adapter (F-G)
  & $0.039{\pm}0.004$ & $0.148{\pm}0.009$ & $0.243{\pm}0.004$
  & $0.438{\pm}0.013$ & $0.611{\pm}0.014$ & $0.863{\pm}0.016$ \\
DeCLIP (Global)
  & $0.077{\pm}0.005$ & $0.277{\pm}0.002$ & $0.370{\pm}0.016$
  & $0.586{\pm}0.004$ & $0.700{\pm}0.014$ & $0.843{\pm}0.011$ \\
DeCLIP (F-G)
  & $0.038{\pm}0.013$ & $0.136{\pm}0.014$ & $0.239{\pm}0.004$
  & $0.400{\pm}0.007$ & $0.566{\pm}0.020$ & $0.754{\pm}0.011$ \\
Global LoRA
  & $0.100{\pm}0.012$ & $0.350{\pm}0.024$ & $0.510{\pm}0.020$
  & $0.737{\pm}0.016$ & $0.882{\pm}0.018$ & $0.960{\pm}0.010$ \\
F-G LoRA
  & $0.048{\pm}0.002$ & $0.185{\pm}0.018$ & $0.282{\pm}0.008$
  & $0.489{\pm}0.005$ & $0.666{\pm}0.019$ & $0.848{\pm}0.036$ \\
T-CLIP (Global)
  & $\underline{0.101{\pm}0.006}$ & $\underline{0.385{\pm}0.012}$ & $\underline{0.556{\pm}0.016}$
  & $\underline{0.787{\pm}0.014}$ & $\underline{0.901{\pm}0.005}$ & $\underline{0.970{\pm}0.006}$ \\
T-CLIP (F-G)
  & $0.039{\pm}0.006$ & $0.144{\pm}0.002$ & $0.219{\pm}0.013$
  & $0.346{\pm}0.020$ & $0.526{\pm}0.022$ & $0.708{\pm}0.006$ \\
\textbf{T-CLIP (Dual)}
  & $\mathbf{0.104{\pm}0.015}$ & $\mathbf{0.395{\pm}0.034}$ & $\mathbf{0.587{\pm}0.009}$
  & $\mathbf{0.818{\pm}0.016}$ & $\mathbf{0.914{\pm}0.003}$ & $\mathbf{0.979{\pm}0.008}$ \\
\rowcolor{blue!6}
$\Delta$ vs.\ Global LoRA
  & $+$4.0\% & $+$12.9\% & $+$15.1\%
  & $+$11.0\% & $+$3.6\% & $+$2.0\% \\
\end{tabular}
}
\end{center}
\end{table}


\subsection{Thermal Image Generation: User Study, Quantitative Metrics, and Additional Samples }
\label{sec:supp_generation}

This appendix provides complete protocol details supporting the user study of generated thermal images presented in 
Figure~\ref{fig:special_images},~\ref{fig:supp_gen_samples} and~\ref{fig:supp_special_condition}.
section~\ref{sec:supp_blinding} describes the user study protocol,
Table~\ref{tab:supp_d2_checklist} gives the full D2 Physics Correctness
checklist, section~\ref{sec:supp_statistics} gives statistical analysis notes, Figure~\ref{fig:supp_fid_clip} reports FID and CLIP Score of the generated thermal images for reference, and Figures~\ref{fig:supp_gen_samples} and~\ref{fig:supp_special_condition} present additional generated samples.

\subsubsection{Study Protocol}
\label{sec:supp_blinding}

Method identity was concealed throughout: annotators saw only
``Image~A'' and ``Image~B'', with A/B labels randomised independently
per annotator per pair using a fixed per-annotator seed to prevent
position bias.
The full text prompts were displayed above both images for each pair.
Three practice pairs (excluded from analysis) preceded the main study.
A private mapping file recorded the A/B assignment per annotator per
pair; post-hoc, each A/B response was converted to a method-level
score as $s^{\text{tclip}}_{a,p,d} = r^A_{a,p,d}$ if Image~A was
T-CLIP, else $r^B_{a,p,d}$, yielding $n{=}150$ paired scores per
dimension (5 annotators $\times$ 30 pairs). 

\subsubsection{D2 Physics Correctness Checklist}
\label{sec:supp_d2}

\paragraph{Scoring formula.}
For each image $i$ and annotator $a$, let
$c_{a,i,j} \in \{0, 1, \text{N/A}\}$ denote the response to
checklist item $j$.
The per-image D2 score is:
\begin{equation}
  s^{D2}_{a,i} =
  \left\lfloor
    \frac{\displaystyle\sum_{j:\,c_{a,i,j} \neq \text{N/A}}
          c_{a,i,j}}
         {|\{j : c_{a,i,j} \neq \text{N/A}\}|}
    \times 3
  \right\rceil + 1,
  \label{eq:d2_score}
\end{equation}
clamped to $[1,4]$, where $\lfloor\cdot\rceil$ denotes rounding to
the nearest integer.
N/A items are excluded from both numerator and denominator.
If all items are N/A the pair is excluded from D2 analysis.\\
Refer to Table~\ref{tab:supp_d2_checklist} for the detailed checklist for physics correctness.

\begin{table}[t]
\caption{Full D2 physics correctness checklist
  (18 items, 7 scene-content groups).
  Fundamental rule: warm\,$\to$\,brighter; cool\,$\to$\,darker.
  Score\,$=$\,proportion YES among applicable items,
  mapped to 1--4 via Eq.~\ref{eq:d2_score}.
  Typical applicable items per image: 6--9.}
\label{tab:supp_d2_checklist}
\begin{center}
\resizebox{\linewidth}{!}{%
\begin{tabular}{l l l l}
\multicolumn{1}{c}{\bf ITEM} &
\multicolumn{1}{c}{\bf NAME} &
\multicolumn{1}{c}{\bf WHAT TO CHECK} &
\multicolumn{1}{c}{\bf YES\,/\,NO CRITERION} \\
\hline \\
\rowcolor{black!10}\multicolumn{4}{l}{\textit{Group A --- Always applicable (no N/A)}} \\
A1 & Warm$\to$brighter; Cool$\to$darker
   & Both directions hold: warm bright AND cool dark
   & YES: bidirectional contrast\newline NO: uniform or inverted \\
A2 & Sky as cold thermal sink
   & Open sky one of darkest regions
   & YES: sky dark\quad NO: sky warm \\
A3 & Three-level hierarchy
   & Sky $<$ Road $<$ People/Vehicles
   & YES: three distinct levels\newline NO: road $=$ sky or road $=$ people \\ \\
\rowcolor{black!10}\multicolumn{4}{l}{\textit{Group B --- People present (N/A if no people)}} \\
B1 & Body heat distribution
   & Torso/head brightest; extremities darker
   & YES: upper body brighter than limbs\newline NO: uniform glow \\
B2 & Clothing insulation
   & Clothed areas slightly darker than bare skin
   & YES: clothed areas dimmer\newline NO: uniform regardless of clothing \\
B3 & Umbrella as cold shield
   & Umbrella dome dark --- blocks body heat
   & YES: umbrella darker than person\newline N/A: no umbrella \\ \\
\rowcolor{black!10}\multicolumn{4}{l}{\textit{Group C --- Vehicles present (N/A if no vehicles)}} \\
C1 & Engine/exhaust heat
   & Engine warm $\to$ brighter than body panels
   & YES: engine brighter\quad NO: uniform \\
C2 & Tyre heat
   & Tyres warm from friction $\to$ brighter than road
   & YES: tyres brighter than road \\
C3 & Moving vs stationary
   & Moving warmer $\to$ brighter
   & YES: moving clearly brighter\newline N/A: only one type present \\
C4 & Headlights as bright spots
   & Headlights $\to$ brightest concentrated spots
   & YES: headlights brightest\newline N/A: no night scene \\  \\
\rowcolor{black!10}\multicolumn{4}{l}{\textit{Group D --- Road visible (N/A if road not visible)}} \\
D1 & Road thermal state
   & Day: warm $\to$ bright; Night: intermediate
   & YES: correct for time of day\newline NO: road as dark as sky \\
D2 & Wet surface emissivity
   & Wet ($\varepsilon{\approx}0.97$) brighter than dry ($\varepsilon{\approx}0.90$)
   & YES: wet areas brighter\quad N/A: no wet surfaces \\  \\
\rowcolor{black!10}\multicolumn{4}{l}{\textit{Group E --- Adverse weather (N/A if clear)}} \\
E1 & Weather reduces contrast
   & Fog/rain $\to$ reduced warm/cool differentiation
   & YES: reduced contrast\quad N/A: clear weather \\
E2 & Overcast sky still dark
   & Overcast: sky dark but less extreme than clear night
   & YES: sky dark but not fully black\newline N/A: not overcast \\ \\
\rowcolor{black!10}\multicolumn{4}{l}{\textit{Group F --- Vegetation present (N/A if no vegetation)}} \\
F1 & Vegetation cooling
   & Evapotranspiration $\to$ vegetation darker than buildings
   & YES: vegetation clearly darker\newline NO: same as warm objects \\  \\
\rowcolor{black!10}\multicolumn{4}{l}{\textit{Group G --- Buildings present (N/A if no buildings)}} \\
G1 & Window-wall differential
   & Windows warm $\to$ brighter; walls cooler $\to$ darker
   & YES: windows brighter than wall\newline NO: building uniformly bright \\
G2 & Rooftop cooler than facade
   & Rooftop $\to$ darker than facade below
   & YES: rooftop darker\quad N/A: rooftop not visible \\ \\ \\
\multicolumn{4}{l}{\textit{Score mapping:
  $0$--$25\%$ YES $\to$ 1;\;
  $26$--$50\%$ $\to$ 2;\;
  $51$--$75\%$ $\to$ 3;\;
  $76$--$100\%$ $\to$ 4.\;
  If all N/A: exclude from D2 analysis.}} \\
\end{tabular}
}
\end{center}
\end{table}

\subsubsection{Statistical Analysis}
\label{sec:supp_statistics}

\paragraph{Statistical tests.}
D1 (Thermal Plausibility) and D2 (Physics Correctness) are assessed
via the sign test rather than the Wilcoxon signed-rank test, because
zero-shot SDXL received a unanimous score of~1 on both dimensions
across all 150 ratings, yielding zero variance; the sign test does not
require non-zero differences ($p{\approx}4.9{\times}10^{-46}$ for
both). D3 (Prompt Faithfulness) is assessed via the Wilcoxon
signed-rank test; the difference was not significant ($p{>}0.05$),
consistent with both models generating correct scene content from the
prompt. Benjamini-Hochberg correction (FDR\,$=$\,0.05) is applied
across all three pairwise comparisons to control the false discovery
rate. Annotator preference (T-CLIP\,+\,SDXL vs.\ zero-shot SDXL) is
assessed via a one-sided binomial test ($H_0$: preference
probability\,$=$\,0.5); all 150 decisive comparisons favoured
T-CLIP\,+\,SDXL, yielding $p{\approx}4.9{\times}10^{-46}$, as
reported in Table 10. of ,aim manuscript.

\paragraph{Score aggregation.}
For D1 and D3, each annotator assigns a Likert score
$s_{a,p,d} \in \{1,2,3,4\}$ per pair.
For D2, scores are derived from a per-image checklist
via equation~\ref{eq:d2_score}.
In all cases, the reported mean and standard deviation are computed
over all $n{=}150$ paired scores (5 annotators $\times$ 30 pairs) as follows:
\begin{equation}
  \mu_d = \frac{1}{150}\sum_{a,p} s_{a,p,d}, \qquad
  \sigma_d = \mathrm{std}\bigl(\{s_{a,p,d}\}_{a,p}\bigr),
\end{equation}
where $s_{a,p,d}$ is the score assigned by annotator $a$ to pair $p$
on dimension $d$. The method-level mean difference on dimension $d$ is
$\Delta = \mu_d^{\text{T-CLIP}} - \mu_d^{\text{SDXL}}$.
\subsubsection{FID and CLIP Score}
\label{sec:supp_fid_clip}

\begin{figure}[h]
\begin{center}

\includegraphics[width=0.55\linewidth]{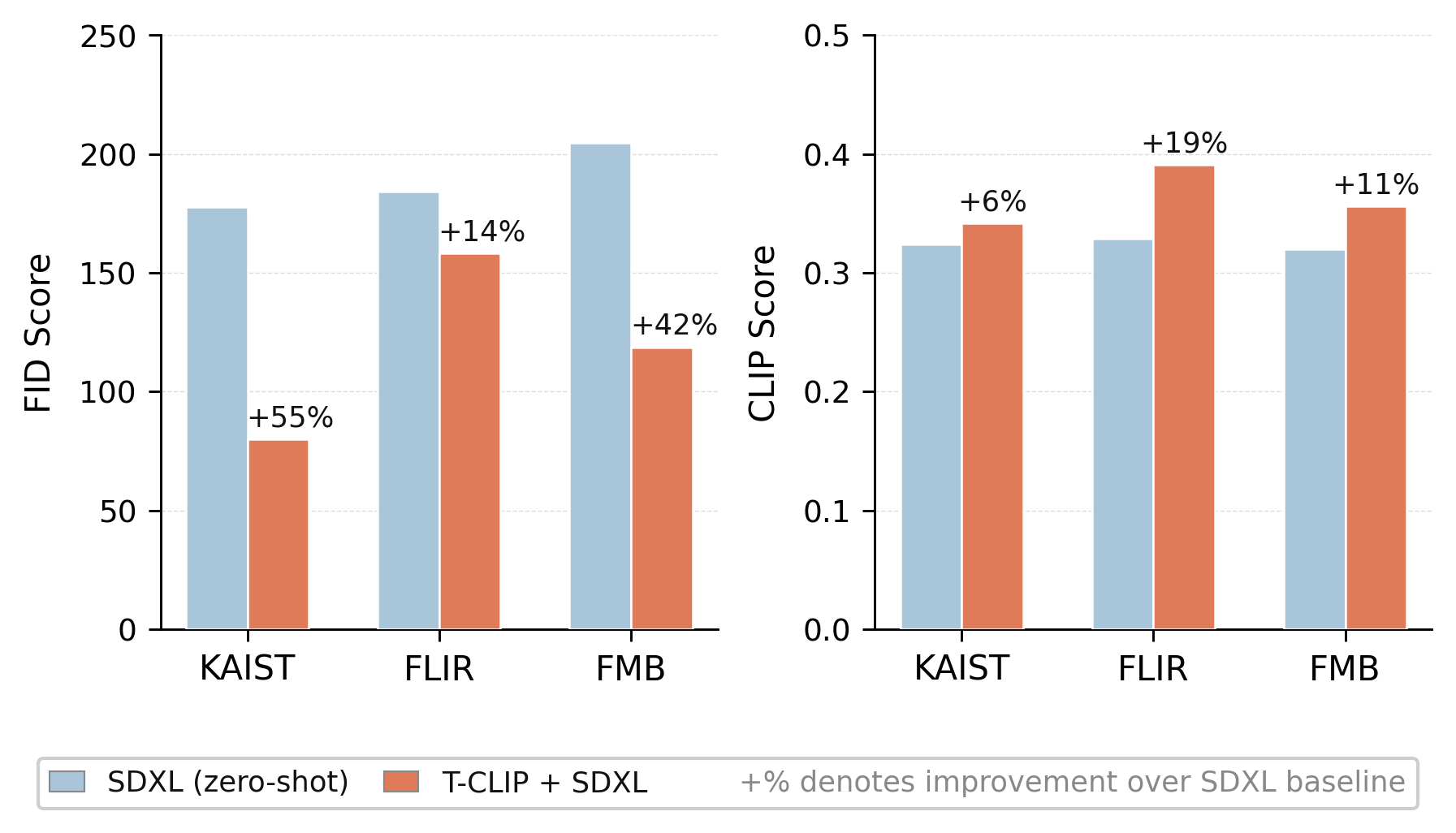}
\end{center}
\caption{FID (lower is better) and CLIP Score (higher is better) for
         zero-shot SDXL vs T-CLIP\,+\,SDXL across
         KAIST~\citep{hwang2015multispectral},
         FLIR~\citep{flir}, and FMB~\citep{liu2023multi}.
         }

\label{fig:supp_fid_clip}
\end{figure}

Figure~\ref{fig:supp_fid_clip} reports FID and CLIP Score for completeness, providing additional quantitative evaluation of the generated thermal images beyond the user study.
Both metrics rely on RGB-trained features and are fundamentally
misaligned with the thermal domain; the user study
(Section~5.6 of the main manuscript) provides the primary evaluation.
The consistent FID improvements ($+14\%$ to $+55\%$) and CLIP Score
improvements ($+6\%$ to $+19\%$) across all three datasets are
nonetheless directionally supportive.

\paragraph{Data Availability.}
All 30 prompts and corresponding generated image pairs used for the user study of thermal image generation are available at
\url{https://drive.google.com/drive/folders/1mxPTpRJgI54XvG5Je8LTEci9PlslPYSK?usp=drive_link}.

\subsubsection{Additional Generated Samples}

\label{sec:gen_samples}
\begin{figure}[h]
\begin{center}

\includegraphics[width=0.95\linewidth]{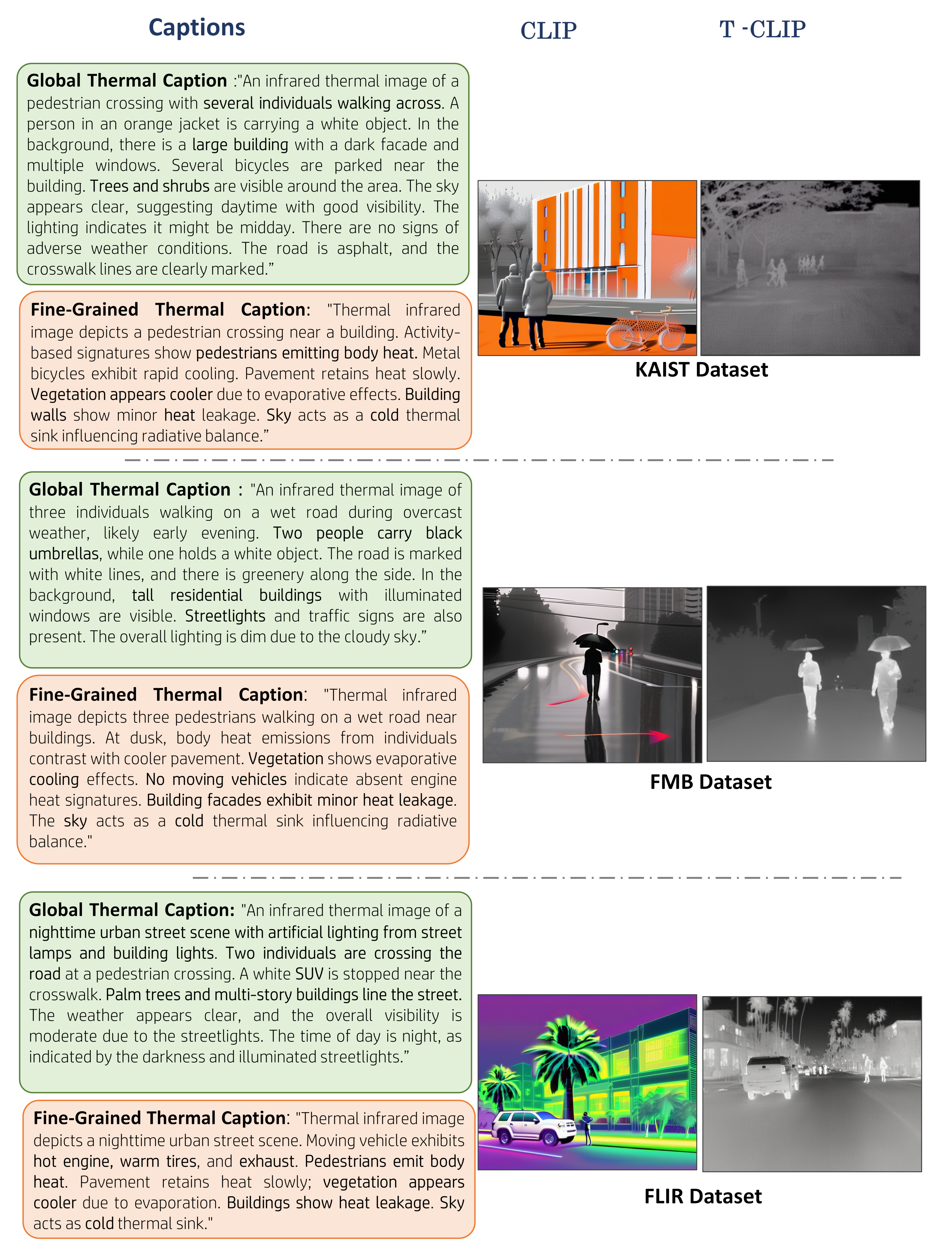}
\end{center}
\caption{Generated thermal image samples using captions from KAIST~\citep{hwang2015multispectral}, FLIR~\citep{flir}, and FMB~\citep{liu2023multi}. For each prompt, left: zero-shot SDXL (standard CLIP text encoder); right: T-CLIP\,+~SDXL. The T-CLIP\,+~SDXL model captures both global scene context and fine-grained object-level heat signatures.}

\label{fig:supp_gen_samples}
\end{figure}

\begin{figure}[h]
\begin{center}
\includegraphics[width=0.95\linewidth]{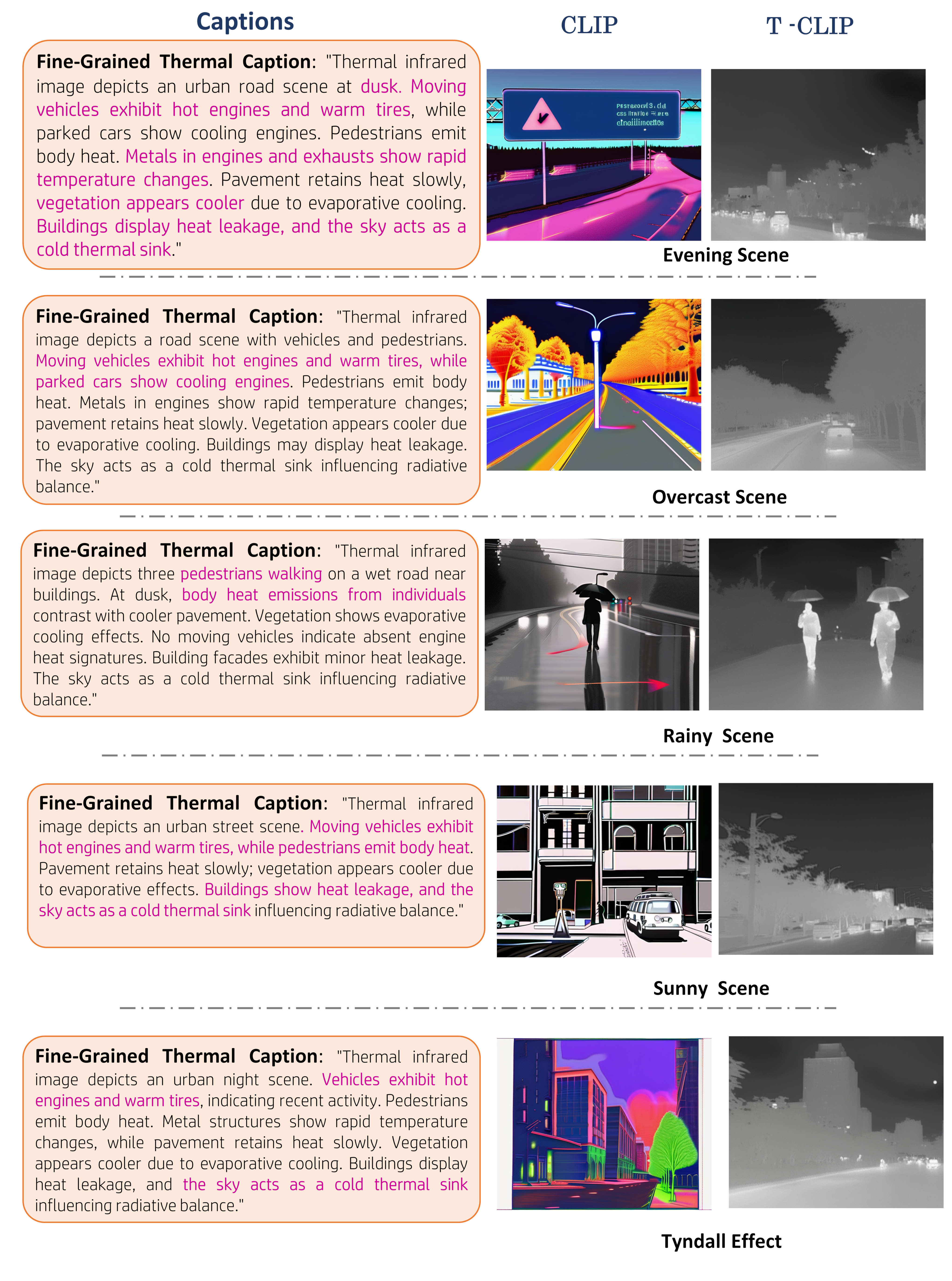}
\end{center}
\caption{T-CLIP\,+~SDXL performance under challenging conditions from FMB \citep{liu2023multi} dataset. For each pair, left: zero-shot SDXL
  (standard CLIP text encoder); right: T-CLIP\,+~SDXL. T-CLIP\,+~SDXL
  demonstrates thermal understanding in scenarios where conventional
  RGB-based approaches fail.}
\label{fig:supp_special_condition}
\end{figure}

We provide additional qualitative examples of thermal images generated using T-CLIP with SDXL using a variety of text prompts in Figures~\ref{fig:supp_gen_samples} and~\ref{fig:supp_special_condition}.


\subsection{Implementation, Training, and Hyperparameter Details}
\label{sec:supp_hyperparams}

All methods use CLIP \citep{clip} ViT-Base-Patch16 as the backbone and are trained
on thermal image--caption pairs generated by the IR-Cap pipeline.
Table~\ref{tab:supp_shared_config} lists hyperparameters shared across all
methods; Table~\ref{tab:supp_method_config} lists method-specific
configurations.
The T-CLIP training and inference pseudocode is shown in
Figure~\ref{fig:suppl_pseudocode}.

\subsubsection{Shared Training Configuration}
\label{sec:supp_shared}

Table~\ref{tab:supp_shared_config} lists the implementation details and shared training configurations used across all experiments. All experiments were performed on a single NVIDIA RTX A6000 GPU with 48\,GB of memory.

\begin{table}[t]
\caption{Shared training configuration across all methods.}
\label{tab:supp_shared_config}
\begin{center}

\begin{tabular}{ll}
\multicolumn{1}{c}{\bf PARAMETER} & \multicolumn{1}{c}{\bf VALUE} \\
\hline \\
Base model          & CLIP ViT-Base-Patch16 \\
Input data          & Thermal image--caption pairs (IR-Cap) \\ 
Optimizer           & AdamW \\
$\beta_1,\,\beta_2$ & 0.9,\;0.999 \\
$\epsilon$          & $1.0\times10^{-8}$ \\
LR scheduler        & Cosine with 100 warmup steps \\
Weight decay        & $1.0\times10^{-3}$ \\
Max gradient norm   & 1.0 \\ 
Train batch size    & 128 \\
Gradient accumulation & 1 \\
Precision           & FP16 \\
Data workers        & 16 \\
GPU                 & NVIDIA A6000 (48\,GB) \\
\end{tabular}
\end{center}
\end{table}

\subsubsection{Method-Specific Configuration}
\label{sec:supp_method_specific}

Table~\ref{tab:supp_method_config} summarises the configuration parameters
unique to each method.
LoRA (Global) and LoRA (F-G) share identical hyperparameters, differing
only in the caption type used for training.
CLIP-Adapter uses a residual bottleneck adapter on top of frozen CLIP
features, controlled by the blend ratio~$\alpha_\text{adp}$.
DeCLIP fine-tunes all CLIP parameters with an additional
image--image and text--text nearest-neighbour consistency loss
computed against a momentum encoder (EMA update), using a lower
learning rate. All methods are trained for an identical number of optimization steps (10{,}000) to ensure a fair comparison.
\begin{figure}[h]
\begin{center}
\includegraphics[width=\linewidth]{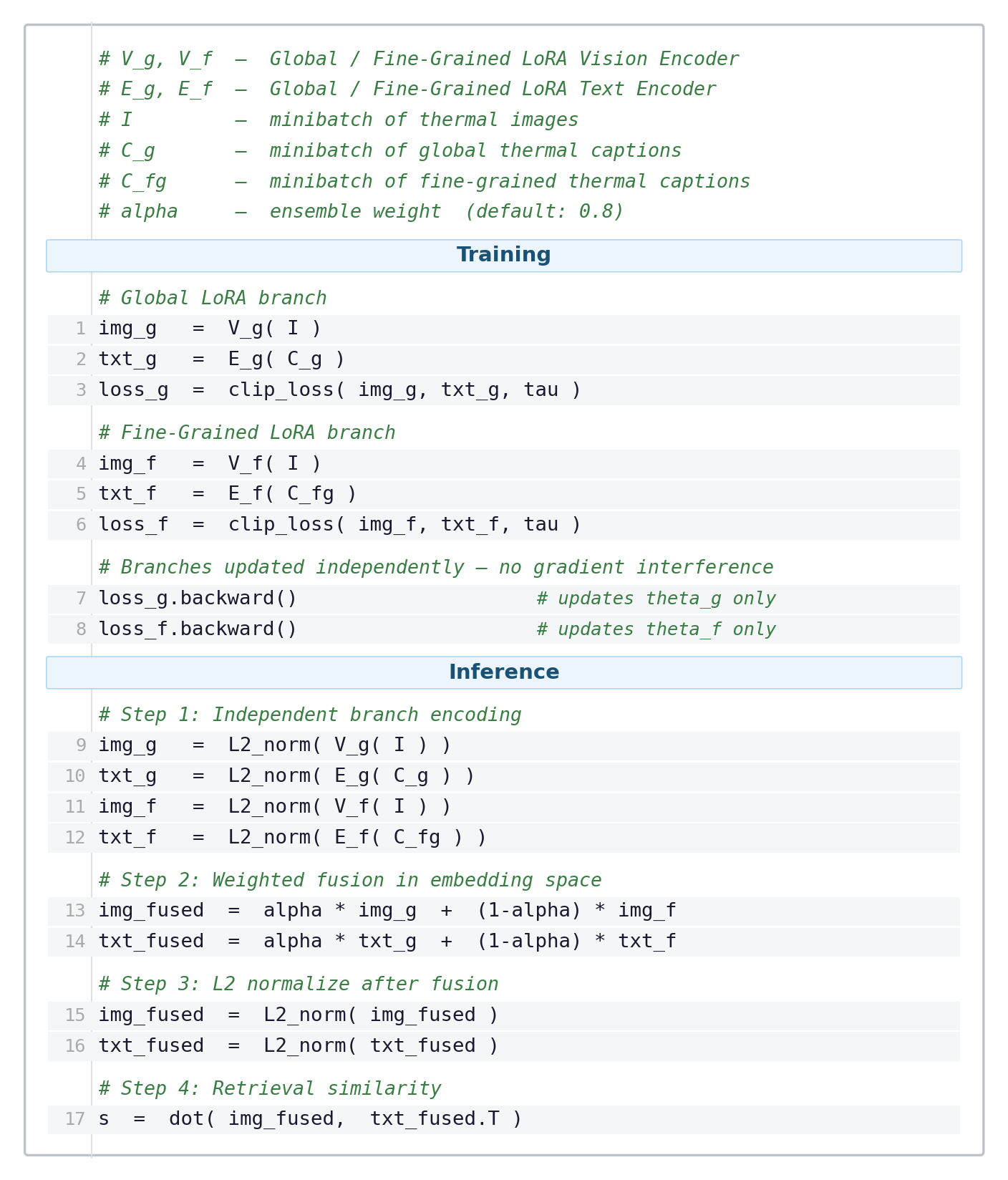}
\end{center}
\caption{T-CLIP training and inference pseudocode.
The global LoRA ($\theta_g$) and fine-grained
LoRA ($\theta_f$) branches are optimised
independently on semantically distinct caption
types with no shared gradient flow, preventing
representational interference.
During inference, L2-normalised features from
both branches are fused via the ensemble weight
$\alpha{=}0.8$ before final similarity
computation (see section~\ref{sec:inference}).}
\label{fig:suppl_pseudocode}
\end{figure}
\begin{table}[t]
\caption{Method-specific hyperparameters.
  LoRA (Global) and LoRA (F-G) share identical settings,
  trained on global and fine-grained captions respectively.
  DeCLIP uses a lower learning rate.
  ``---'' denotes not applicable.}
\label{tab:supp_method_config}
\begin{center}
\resizebox{\linewidth}{!}{%
\begin{tabular}{llll}
\multicolumn{1}{c}{\bf PARAMETER} &
\multicolumn{1}{c}{\bf LORA (GLOBAL / F-G)} &
\multicolumn{1}{c}{\bf CLIP-ADAPTER~\citep{gao2024clip}} &
\multicolumn{1}{c}{\bf DECLIP~\citep{declip}} \\
\hline \\
Trainable parameters  & LoRA adapters only        & Bottleneck MLP only       & All CLIP parameters \\
Frozen backbone       & \ding{51}                 & \ding{51}                 & \ding{55} \\
LoRA rank $r$         & 8                         & ---                       & --- \\
LoRA $\alpha$         & 1                         & ---                       & --- \\
LoRA dropout          & 0                         & ---                       & --- \\
Target modules        & \texttt{q,k,v\_proj}      & ---                       & --- \\
Adapter blend ratio   & ---                       & 0.2                       & --- \\
Adapter bottleneck    & ---                       & $512{\to}128{\to}512$     & --- \\
\\
Learning rate         & $2.0\times10^{-3}$        & $1.0\times10^{-3}$        & $1.0\times10^{-5}$ \\
Max steps             & 10,000                    & 10,000                    & 10,000 \\
Training time         & ${\approx}1.5$\,hrs       & ${\approx}40$\,min        & ${\approx}2$\,hrs \\
\\
$\lambda_\text{ii}$ (image-image weight) & --- & --- & 0.5 \\
$\lambda_\text{tt}$ (text-text weight)   & --- & --- & 0.5 \\
NN temperature $\tau$ & ---                       & ---                       & 0.1 \\
Top-$k$ NNs           & ---                       & ---                       & 1 \\
Momentum $m$ (EMA)    & ---                       & ---                       & 0.995 \\
Memory bank size      & ---                       & ---                       & 4,096 \\ 
\\
\rowcolor{black!10}\multicolumn{4}{p{18cm}}{LoRA target modules: \texttt{q\_proj},
\texttt{k\_proj}, \texttt{v\_proj} in both vision and text encoders.
CLIP-Adapter blend: output\,$=$\,$0.2{\times}\text{adapter}(x)+0.8{\times}x$.
DeCLIP memory bank stores the last 4,096 momentum encoder features in a
FIFO queue; NNs searched over the combined bank and current batch for
stable target computation.} \\
\end{tabular}
}
\end{center}
\end{table}

\subsubsection{T-CLIP Training and Inference Pseudocode}
\label{sec:supp_pseudocode}

Figure~\ref{fig:suppl_pseudocode} illustrates the T-CLIP training and
inference procedure.
The Global LoRA ($\theta_g$) and Fine-Grained LoRA ($\theta_f$)
branches are optimised independently on their respective caption types
with no shared gradient flow, and fused at inference via ensemble
weight $\alpha{=}0.8$.

\end{document}